\documentclass[preprint,12pt,authoryear]{elsarticle}

\usepackage[T1]{fontenc}
\usepackage{lmodern}
\usepackage{amsmath}
\usepackage{amssymb}
\usepackage{booktabs}
\usepackage{graphicx}
\usepackage{multirow}
\usepackage{makecell}
\usepackage{array}
\usepackage{xcolor}
\usepackage{microtype}
\usepackage[colorlinks=true,allcolors=blue!60!black]{hyperref}
\hypersetup{pdftitle={Machine learning for journal entry testing: A type-aware evaluation of anomaly detectors under a review budget},
  pdfauthor={Jan Gronewald, Michel Scherer, Nijat Mehdiyev}}
\journal{Expert Systems with Applications}

\begin{document}

\begin{frontmatter}

\title{Machine learning for journal entry testing: A type-aware evaluation of anomaly detectors under a review budget}

\author[uds,dfki]{Jan Gronewald\corref{cor1}}
\ead{jan.gronewald@extern.uni-saarland.de}
\author[uds,dfki]{Michel Scherer}
\author[leipzig,dfki]{Nijat Mehdiyev}

\cortext[cor1]{Corresponding author.}

\affiliation[uds]{organization={Saarland University},
            addressline={Campus},
            postcode={66123},
            city={Saarbr\"ucken},
            country={Germany}}

\affiliation[dfki]{organization={German Research Center for Artificial Intelligence (DFKI), Institute for Information Systems (IWi)},
            addressline={Campus D3.2},
            postcode={66123},
            city={Saarbr\"ucken},
            country={Germany}}

\affiliation[leipzig]{organization={Leipzig University},
            addressline={Grimmaische Straße 12},
            postcode={04109},
            city={Leipzig},
            country={Germany}}

\begin{abstract}
Journal entry anomaly detectors are commonly evaluated on the full population with ROC-AUC, precision and recall, ignoring the review budget and which anomaly types are found. We propose a type-aware evaluation combining per-type recall, fair-share type recall (FSR), which caps each type's credit at its budget share, type coverage and first-hit rank. We evaluate nine unsupervised detectors, a supervised reference and feedback-driven Deep Semi-Supervised Anomaly Detection (DeepSAD) on four real client ledgers with injected typed anomalies and a public synthetic ledger. On the largest client ledger, principal component analysis (PCA), an autoencoder (AE) and a variational autoencoder (VAE) each place on average 98 anomalies among the first 100 postings, but at least 95.8 belong to one type. FSR instead favours a nearest-neighbour (kNN) detector and changes the top-ranked detector on three of four client ledgers. Representation also matters: one-hot encoding exposes unseen accounts, whereas frequency encoding leaves unseen contra accounts largely undetected. On the public ledger, the Histogram-Based Outlier Score (HBOS) and Empirical Cumulative Distribution-Based Outlier Detection (ECOD) reach all eight markings within 1,386 entries, whereas kNN, the hit leader at 1,000 entries, first reaches cross-linked clearing at rank 4,641, and the supervised row-level reference misses this marking within 1,000 entries. There, the adaptive DeepSAD review protocol raises mean hits per 100 reviews from 40.0 to 68.3 but type coverage only from 2.7 to 3.0. These findings show that high hit rates can conceal systematic blind spots and suggest that feedback can reinforce existing detection patterns without broadening anomaly coverage.
\end{abstract}


\end{frontmatter}

\section{Introduction}\label{sec:intro}

Journal entry testing is among the most direct applications of machine learning in auditing. International Standard on Auditing (ISA) 240 requires the auditor to test the appropriateness of journal entries recorded in the general ledger \citep{iaasb2025isa240}, yet a single client's ledger for one fiscal year holds thousands to hundreds of thousands of postings, of which an engagement team can examine only a few in detail. Practice therefore selects entries by risk criteria \citep{droste2024journal}, and guidance from the International Auditing and Assurance Standards Board (IAASB) accepts that auditors follow up only a selection of the items that automated tools flag \citep{iaasb2023technology}. Because genuine misstatements are rare and almost never labelled, the natural tool is an unsupervised anomaly detector that learns the regularities of the postings, scores every entry and hands the auditor a ranking to work through from the top. The result is an expert system in which the model proposes and the auditor disposes, and its value in practice is decided by the review budget, the number $k$ of entries that the auditor actually reviews. Evaluating detectors for this setting therefore means asking what the first $k$ entries of a ranking contain.

The measures with which machine learning compares anomaly detectors answer this question only in part. Precision at $k$ ($\mathrm{P@}k$), average precision (AP) and the area under the receiver operating characteristic curve (ROC-AUC) measure how well a ranking places anomalies ahead of normal instances, but they treat anomalies as exchangeable and ignore which kinds they are. Benchmark studies have shown that this matters: the ranking of algorithms changes with the anomaly type, because each algorithm carries an inductive bias, a built-in assumption about what makes a point anomalous, that fits some deviations but not others \citep{steinbuss2021benchmarking,han2022adbench,foorthuis2021nature}. These studies, however, evaluate one type per dataset. In a general ledger the types coexist in one ranking, and for the auditor they are not interchangeable: an extreme amount, an unusual payment period, a posting to a never-used account and an unusual pair of accounts point to different risks, and ISA 330 asks for procedures that respond to each assessed risk \citep{iaasb2025isa330}. A type-blind measure therefore rewards a model whose inductive bias happens to match the most numerous or most separable type, and it cannot reveal that the review list leaves the other risks untested.

A single example from our benchmark shows how far the two criteria can drift apart. In a real client ledger of about 280,000 postings with five injected anomaly types, principal component analysis (PCA), an autoencoder (AE) and a variational autoencoder (VAE) each place on average 98 anomalies among their first 100 postings, a nearly perfect result by any hit-based measure. Yet at least 95.8 of these hits are of one kind, postings to an account never used before. We attribute this to the interplay of inductive bias and representation: the three models score a posting by how poorly it is reconstructed from structure learned from the data, and in a one-hot encoding an unused account becomes a column of its own that no learned structure explains. A measure that credits each type only up to its fair share of the budget, fair-share type recall ($\mathrm{FSR@}k$, Section~\ref{sec:evaluation}), reverses the verdict: a k-nearest-neighbour detector (kNN) that finds only a third as many anomalies but spreads them over several types scores higher ($\mathrm{FSR@}100$ of at least 0.284 against at most 0.226).

\begin{figure}[tbp]
\centering
\includegraphics[width=\textwidth]{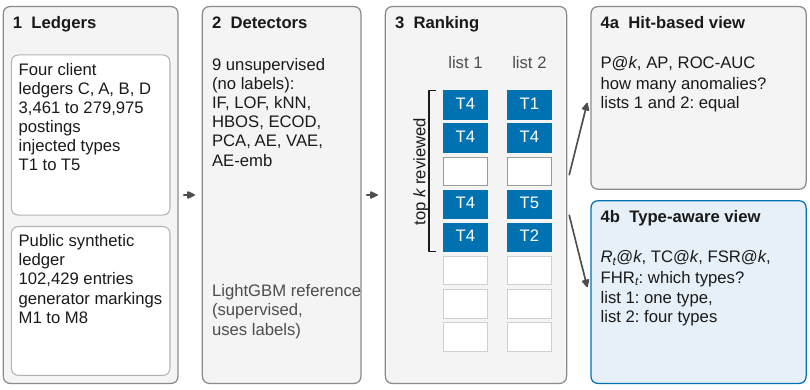}
\caption{Type-aware budget evaluation. Four client ledgers with injected types T1 to T5 and a public ledger with markings M1 to M8 are ranked by nine unsupervised detectors and a supervised LightGBM reference; an auditor reviews the first $k$ units. The conventional view counts the anomalies among them, the type-aware view asks which types they are. The example lists are schematic.}
\label{fig:framework}
\end{figure}

This paper develops a type-aware evaluation for anomaly detection under a review budget and tests whether it changes the conclusions of a hit-based comparison. The evaluation consists of per-type recall read against the most the budget can hold, type coverage, $\mathrm{FSR@}k$ with optional risk weights, the first-hit rank of each type and a budget scaled to ledger size. The measures need only model scores and anomaly types, so they apply to any detector. They adapt diversity-aware evaluation of ranked lists from information retrieval to three features of the audit setting: the expert reads the whole list, the types differ greatly in size, and their weights come from a risk assessment rather than from the probability of a query intent (Fig.~\ref{fig:framework}).

We apply the evaluation to three learning regimes on two complementary data sources. Typed synthetic anomalies of five kinds are injected into four real client ledgers of 3,461 to 279,975 postings under a protocol frozen before the experiments, and a public synthetic ledger of 102,429 journal entries carries eight markings defined by its generator \citep{gronewald2026generating}, independently of our injection protocol and open to anyone. The unsupervised regime comprises nine detectors whose inductive biases range from isolation and local density to linear subspaces, neural reconstruction and learned entity embeddings. In the supervised regime, gradient-boosted trees trained on the labels serve as a supervised reference that indicates which types the row-level representation makes separable at all. In the third regime the auditor's reviews become labels for a semi-supervised deep one-class model, DeepSAD \citep{ruff2020deep}. Such labels come from the top of the model's own ranking, a sample known to be unrepresentative of the data \citep{dasgupta2008hierarchical,pelleg2004active}, which raises the question whether learning from the expert widens the range of risks reached or only deepens the one already found. Five research questions structure the study.
\begin{description}
\item[RQ1] How do the detectors compare by conventional, hit-based criteria under a review budget?
\item[RQ2] Which anomaly types reach the reviewer within the budget, and does the type-aware view change the comparison?
\item[RQ3] How do budget size, contamination and difficulty move both views?
\item[RQ4] Which findings hold on a public ledger whose anomalies were generated independently of this benchmark?
\item[RQ5] Do auditor reviews fed back to a semi-supervised detector reach more anomaly types, or only more anomalies?
\end{description}

The answers show that the two readings disagree where it matters. On three of the four client ledgers the model with the highest $\mathrm{FSR@}100$ is not the one with the most hits, and on the largest this reversal requires the fair-share normalisation: macro-averaged recall would keep the hit leaders on top. On the public ledger FSR orders the detectors as precision does, but coverage and first-hit ranks do not: two per-field detectors with low precision reach all eight markings within 1,400 entries, whereas kNN, the leader in hits at a budget of 1,000 entries, first meets a cross-linked clearing at rank 4,641. Relational anomalies almost never enter the first 1,000 entries of any unsupervised detector, and the supervised learner misses cross-linked clearings as well, which points to a limit of the row-level representation rather than of the model. The adaptive review protocol with DeepSAD adds about 17 to 49 anomalies per 100 reviews over its static starting ranking. On the public ledger a control without labels attributes this gain on average to the auditor's labels, yet the labelled hits concentrate on one marking per seed and reach no more markings than retraining without labels, evidence of an exploitation bias in ranking-derived feedback.

The paper makes three contributions: a type-aware evaluation framework for budget-constrained anomaly detection in human-in-the-loop decision support, including the fair-share normalisation required when anomaly categories of very unequal size coexist in one ranking; a benchmark that combines typed injection into four real client ledgers with a public synthetic ledger and compares unsupervised, supervised and expert-feedback learning; and evidence, together with the mechanisms behind it, that the type-aware view changes which model would be selected under a type-aware criterion. The detectors and the injection method are established; the new elements are the fair-share normalisation, its combination with coverage and first-hit ranks into a budget-bound evaluation profile, and the evidence that this profile changes which model would be selected. Section~\ref{sec:related} reviews related work, Section~\ref{sec:evaluation} defines the evaluation, Section~\ref{sec:benchmark} describes the benchmark, Section~\ref{sec:results} reports the results, Section~\ref{sec:discussion} discusses mechanisms, implications and limitations, and Section~\ref{sec:conclusion} concludes.

\section{Related work}\label{sec:related}

Four lines of work meet in this study: the evaluation of anomaly detectors, machine learning for journal entry testing, learning from expert feedback and diversity-aware evaluation of ranked lists. We review each with the question of what it contributes to, and leaves open for, an evaluation of detectors under a review budget.

\subsection{Evaluating anomaly detectors}\label{sec:related-eval}

Unsupervised anomaly detection comprises families that differ in what they take an anomaly to be: a point that is easy to isolate, that lies in a region of low density, far from its neighbours or in the tail of a marginal distribution, or that a model of the normal data reconstructs poorly \citep{chandola2009anomaly,aggarwal2017outlier,pang2021deep}. Benchmark research has shown that these inductive biases decide which anomalies a detector finds. \citet{steinbuss2021benchmarking} generate regular instances from a model of real data and outliers as characterisable deviations from it, with separate local, dependency and global instantiations, and find that the ranking of detectors differs between these types. ADBench \citep{han2022adbench} compares 30 algorithms on 57 datasets and finds no unsupervised algorithm statistically better than the others, but a clear dependence on the match between an algorithm's assumptions and the anomaly type, with the local outlier factor (LOF) best on local and a nearest-neighbour detector best on global anomalies. \citet{foorthuis2021nature} characterises anomalies by the data types and the cardinality of the attributes involved and argues that no single algorithm can identify all types; we use his typology to classify our injected types. For comparing many detectors across many settings, mean ranks with a Friedman test and a Nemenyi post-hoc test are standard \citep{demsar2006statistical}. Yet in the type-specific experiments of these benchmarks each dataset holds one type, and each ranking is scored as a whole, by ROC-AUC or the area under the precision-recall curve; precision at $n$, the measure closest to a review budget, appears among the measures of outlier benchmarks \citep{campos2016evaluation} but likewise counts hits without distinguishing them. When several types coexist in one ranking, as in a ledger, the benchmarks do not ask which of them a fixed list contains.

\subsection{Machine learning for journal entry testing}\label{sec:related-jet}

Detectors for accounting data have developed from data mining to deep and language-model representations. Early work flagged suspicious general ledger accounts across thousands of clients by combining outlier detection with classification \citep{bay2006large}, tested the digit distributions of journal entries against Benford's law \citep{debreceny2010data} and clustered procurement transactions of a single company to reduce internal fraud risk \citep{jans2010internal}.
\citet{schreyer2017detection} trained deep autoencoders on journal entries from an enterprise resource planning (ERP) system and evaluated them on injected global and local anomalies, and \citet{schreyer2019detection} used adversarial autoencoders to separate anomalous entries in the latent space. \citet{schultz2020autoencoder} compared an autoencoder with the judgement of external auditors, and \citet{nonnenmacher2021unsupervised} reviewed unsupervised anomaly detection for internal auditing. Recent studies combine supervised and unsupervised detectors for known and unknown anomalies in general ledger data \citep{bakumenko2022detecting}, encode financial records with large language models \citep{bakumenko2025advancing} and evaluate outlier detection methods on journal entries in a use case analysis \citep{schreier2026evaluation}. The field has thus adopted the models of general anomaly detection together with its evaluation habits \citep{gronewald2026slr}: studies typically use the data of one organisation, inject a few kinds of anomaly, most often global and local ones \citep{schreyer2017detection,schreyer2019detection}, and score whole rankings. To our knowledge none asks which kinds of anomaly a fixed review list contains. A public benchmark ledger now allows such questions to be asked on shared data: \citet{gronewald2026generating} describe a generator of synthetic journal entries and release a general ledger with eight anomaly markings \citep{gronewald2026dib}, six of them injected into the populations that journal entry testing rules select, and \citet{gronewald2026anomaly} report a hybrid approach on general ledger data. The first author of the present paper co-authored these works; the markings were defined without reference to this benchmark, but the data do not come from an independent group.

\subsection{Learning from expert feedback}\label{sec:related-feedback}

When a few labels are available, anomaly detection can be posed as a semi-supervised problem. Deep one-class classification learns a network that maps normal data close to a centre in latent space \citep{ruff2018deep}, DeepSAD extends it to a few labelled normal instances and anomalies, pushing the labelled anomalies away from that centre \citep{ruff2020deep}, and deviation networks train a scoring network from a few labelled anomalies \citep{pang2019deep}. These methods assume that the labels are given in advance, whereas in an audit they arrive sequentially from the reviews themselves. Active anomaly discovery addresses this sequential case by re-weighting an ensemble of detectors with analyst feedback on the top-ranked instances \citep{das2016incorporating}. The active learning literature \citep{settles2009active} records a property of such feedback that matters here: a training set collected by querying selected points diverges from the underlying distribution and is not a representative sample \citep{dasgupta2008hierarchical}, and in anomaly detection labels taken from the top of a ranking concentrate on the points that the ranking already puts first \citep{pelleg2004active}. Whether learning from such labels widens the range of anomaly types that a review list reaches, or only the number of anomalies of the types already found, has not, to our knowledge, been measured.

\subsection{Diversity- and coverage-aware evaluation}\label{sec:related-diversity}

Information retrieval has long evaluated whether a ranked list serves the different intents behind a query. Cumulated gain adds up graded relevance down a list, discounts it by rank and normalises it by an ideal ranking \citep{jarvelin2002cumulated}; subtopic recall at rank $K$ is the share of subtopics covered by the first $K$ documents \citep{zhai2003beyond}; $\alpha$-normalised discounted cumulative gain ($\alpha$-nDCG) lowers the gain of a document for each earlier document on the same intent \citep{clarke2008novelty}; and intent-aware measures average a standard measure over the intents of a query, weighted by their probabilities \citep{agrawal2009diversifying}. The analogy to journal entry testing is close, with anomaly types in the role of intents, but three features of the audit call for an adaptation. The auditor reads the whole list, so what the budget contains matters more than the order within it; the types differ greatly in size, so a type's recall must be read against what the budget can hold and its credit capped at a fair share; and the weights of the types come from a risk assessment, not from the probability of an intent. We know of no measure in anomaly detection that counts which kinds of anomaly reach a fixed budget in a mixed ranking. The detectors, the injection method, the typology and the statistical tests in this study are therefore taken from the work reviewed above, and our contribution is a type-aware evaluation framework for budget-constrained anomaly detection, including the fair-share normalisation that is needed when anomaly categories of very unequal size coexist in one ranking, its application to three learning regimes and the evidence that it changes model-selection conclusions.

\section{Type-aware budget evaluation}\label{sec:evaluation}

The evaluation formalises the auditor's question of what a review list contains. It is model-agnostic: it needs only the scores of a detector and the types of the labelled anomalies, so it applies equally to unsupervised, supervised and feedback-driven models, and it is an evaluation criterion rather than a training objective. It adds four measures and a scaling rule to the hit-based view. We define them in turn, explain the design choices behind them, relate them to existing measures, show how they behave when results are aggregated over seeds and describe how a type profile is read; \ref{app:notation} collects the notation.

\subsection{Hits, types and the budget ceiling}\label{sec:eval-hits}

A ledger holds $N$ units, indexed by their position $i\in\{1,\dots,N\}$ in the file: postings on the client ledgers, journal entries on the public ledger. Each unit has a label $y_i\in\{0,1\}$, and each anomaly has exactly one type $t$ from a set $\mathcal{T}$ of $T=|\mathcal{T}|$ types (T1 to T5, or markings M1 to M8); $\mathcal{I}_t$ holds the anomalies of type $t$, and $n_t=|\mathcal{I}_t|\geq 1$ is their number. A detector gives every unit a score $s_i$, higher meaning more anomalous. The ranking $\pi$ lists the units by descending score, ties broken by ascending position, and $\pi(r)$ is the unit at rank $r$. The auditor reviews $\operatorname{top}_k=\{\pi(1),\dots,\pi(k)\}$, and $f_t(k)=|\operatorname{top}_k\cap\mathcal{I}_t|$ counts the anomalies of type $t$ among them. A type is a labelled anomaly category, defined by the mechanism that produced the anomaly on the client ledgers (for example an unseen account) and by the generator's marking on the public ledger; it is not by itself an audit risk. The measures therefore quantify which labelled categories reach the reviewer, and they measure risk coverage where the categories correspond to the risk classes of an engagement, which is the only sense in which we speak of risks. Precision at $k$, the share of anomalies among the reviewed units \citep{craswell2009precision}, counts hits, and per-type recall at $k$, the share of the anomalies of type $t$ that the auditor sees, shows where they come from:
\begin{equation}
\mathrm{P@}k=\frac{1}{k}\sum_{t\in\mathcal{T}}f_t(k),\qquad
R_t\mathrm{@}k=\frac{f_t(k)}{n_t}\;\leq\;\min\Bigl(1,\frac{k}{n_t}\Bigr).
\label{eq:pk-recall}
\end{equation}
We keep $k$ in the denominator of $\mathrm{P@}k$ even when a ledger holds fewer than $k$ anomalies, because the budget, not the number of anomalies, is what the auditor spends; AP and ROC-AUC complete the conventional view. Every measure has a natural chance level: under a random ordering, $f_t(k)$ follows a hypergeometric distribution with mean $k\,n_t/N$, so the expected precision is the share of anomalies in the ledger, 0.015 on the client ledgers in the core grid.

The bound in Eq.~\eqref{eq:pk-recall} is the budget ceiling. At most $k$ of the $n_t$ anomalies of a type fit into the budget, so a per-type recall must be read against its ceiling rather than against 1: a recall of 0.29 is poor for a type that the budget could hold entirely, but close to the maximum for a type of 336 anomalies at $k=100$. The vector $(R_t\mathrm{@}k)_{t\in\mathcal{T}}$, read against its ceilings, is the type profile of a detector at budget $k$. The simplest summary of this profile is type coverage at $k$, the number of types that the auditor sees at least once,
\begin{equation}
\mathrm{TC@}k=\sum_{t\in\mathcal{T}}\mathbb{I}\bigl[f_t(k)\geq 1\bigr]\in\{0,1,\dots,T\},
\label{eq:tc}
\end{equation}
where $\mathbb{I}[\cdot]$ is the indicator function. Coverage is deliberately coarse, since one hit and a hundred hits of a type count the same; it answers whether a category was touched at all, not how thoroughly, and it therefore needs a companion that rewards depth within each type without rewarding concentration on one.

\subsection{Fair-share type recall}\label{sec:eval-fsr}

A single number that summarises the type profile must decide how much of the budget each type deserves. Shares proportional to the type sizes would reward a list that concentrates on the largest type, which is exactly the behaviour the evaluation should expose; we therefore share the budget equally among the types and redistribute what a small type cannot use. Under this water-filling rule, a type with fewer anomalies than its equal share keeps all of them, and the remainder is split equally among the other types. The cap of type $t$ is
\begin{equation}
c_t=\min(n_t,s^{*}),\qquad\text{where}\ \ \sum_{t\in\mathcal{T}}c_t=\min\Bigl(k,\sum_{t\in\mathcal{T}}n_t\Bigr),
\label{eq:caps}
\end{equation}
and the water level $s^{*}$ is the smallest $s\geq 0$ for which $c_t=\min(n_t,s)$ meets the second condition. Fair-share type recall at $k$ credits each type for its hits up to its cap:
\begin{equation}
\mathrm{FSR@}k=\sum_{t\in\mathcal{T}}w_t\,\frac{\min\bigl(f_t(k),c_t\bigr)}{c_t},\qquad w_t>0,\quad\sum_{t\in\mathcal{T}}w_t=1.
\label{eq:fsr}
\end{equation}
All results in this paper use equal weights, $w_t=1/T$, which should be read as a risk-neutral benchmarking convention rather than a claim that the types matter equally for an audit; Section~\ref{sec:res-robust} and \ref{app:weights} show how the comparisons respond to other weights. A small example shows how the caps work. With a budget of $k=100$ and four types of 10, 30, 200 and 400 anomalies, the equal share is 25; the first type keeps its 10, the remaining 90 units are split among the other three, the second type keeps its 30, and the last two receive 30 each, so the caps are 10, 30, 30 and 30. A list that spends all 100 reviews on the largest type then scores 0.25, whereas a list that fills every cap scores 1, although both may hold 100 hits.

$\mathrm{FSR@}k$ lies in $[0,1]$ and equals 1 exactly when every type receives at least its fair share. It is non-decreasing in every $f_t(k)$, so an additional hit never lowers it, and hits beyond a type's share earn nothing, so a list that fills the budget with one type $t$ scores at most $w_t$, however many hits it holds. We chose a hard cap rather than the geometric decay of $\alpha$-nDCG \citep{clarke2008novelty} because it keeps the measure interpretable as the weighted share of the fair budget that a list fills, and because an auditor who has reviewed a type's fair share has spent the budget allotted to that risk. Two special cases clarify the meaning. When the budget covers all anomalies, as on ledgers C and A at $k=100$, every cap equals $n_t$ and $\mathrm{FSR@}k$ is the weighted mean per-type recall. When every type holds at least $k/T$ anomalies, as on ledgers B and D at $k=100$ and on the public ledger for $k\leq 160$, every cap equals $k/T$; with equal weights $\mathrm{FSR@}k$ is then the precision that remains after discarding hits beyond $k/T$ per type, so it cannot exceed $\mathrm{P@}k$, and the gap between the two measures is the share of the budget spent on surplus hits.

\subsection{First-hit rank and a size-scaled budget}\label{sec:eval-fhr}

Coverage and FSR are read at one budget; the first-hit rank tells the auditor how far to read before each type appears. The first-hit rank of type $t$ is the number of units reviewed until its first anomaly:
\begin{equation}
\mathrm{FHR}_t=\min\bigl\{r:\ \pi(r)\in\mathcal{I}_t\bigr\}.
\label{eq:fhr}
\end{equation}
A type is covered at budget $k$ exactly when $\mathrm{FHR}_t\leq k$, so $\mathrm{TC@}k=\sum_t\mathbb{I}[\mathrm{FHR}_t\leq k]$ is the coverage curve of a ranking, and its maximum, $\max_t\mathrm{FHR}_t$, is the full-coverage budget, the smallest budget at which every type has been seen. First-hit ranks thus summarise the whole coverage curve in $T$ numbers, but they require the full ranking, which is available for the public ledger; for the client ledgers the retained results hold per-type recalls at $k=100$ only. A fixed budget, finally, means different things on ledgers of different size: 100 postings are 2.9 percent of ledger C but 0.04 percent of ledger D. We therefore also use a budget proportional to ledger size,
\begin{equation}
k=\rho N,
\label{eq:budget}
\end{equation}
rounded to a convenient value; on the public ledger, $\rho$ of about 1 percent gives $k=1{,}000$.

\subsection{Relation to existing measures}\label{sec:eval-compare}

FSR adapts the diversity-aware evaluation reviewed in Section~\ref{sec:related-diversity}, so it is fair to ask why an existing measure would not serve. Table~\ref{tab:measures} compares the candidates on the properties that the audit setting requires, and three arguments settle the question. First, applying the intent-aware construction of \citet{agrawal2009diversifying} to precision averages the precision for each intent, weighted by the intent's probability. With disjoint anomaly types the precision for type $t$ at $k$ is $f_t(k)/k$, so the intent-aware precision equals $\sum_t w_t f_t(k)/k$ and, with equal weights, $\mathrm{P@}k/T$; it orders every pair of lists exactly as $\mathrm{P@}k$ does and cannot see concentration.

Second, macro-averaged recall, the mean of $R_t\mathrm{@}k$ over the types, does reward breadth, but it values a hit of type $t$ at $1/(T n_t)$, inversely to the size of the type. When $k<\sum_t n_t$, the list that maximises macro recall therefore spends its budget on the smallest types first, the mirror image of the bias of $\mathrm{P@}k$, and its attainable maximum depends on $k$ and on the type sizes, so its values are not comparable across budgets or ledgers. FSR values a hit of type $t$ at $w_t/c_t$ up to the cap and at zero beyond it; because the caps sum to $\min(k,\sum_t n_t)$ and never exceed $n_t$, every budget admits a list with $\mathrm{FSR@}k=1$, and a list attains it exactly when it fills every fair share. Third, $\alpha$-nDCG \citep{clarke2008novelty} discounts the $j$-th item of a type by $(1-\alpha)^{j-1}$ regardless of the size of the type, so that with $\alpha=0.5$ the fifth unseen account earns a sixteenth of the credit of the first even when the fair share of that type is 20, and it adds a discount by position that an auditor who reads the whole list does not apply. Type coverage, the count form of subtopic recall \citep{zhai2003beyond}, is bound to the budget and needs no size adjustment, but it is binary. FSR is therefore not a replacement for these measures but a member of the family that combines a budget, an adjustment for type size, a cap on over-representation and risk weights; where the order within the budget matters, the first-hit ranks supply it. Where the budget covers every anomaly, the caps equal the type sizes and FSR reduces to macro recall, so its distinctive contribution appears only when the caps are active, as Section~\ref{sec:res-typeaware} shows empirically.

\begin{table}[tbp]
\caption{Properties of hit-based and type-aware measures for a review budget of $k$ units with anomaly types of unequal size. Intent-aware precision is precision under the intent-aware construction of \citet{agrawal2009diversifying}, with anomaly types as intents; macro recall is the mean of $R_t\mathrm{@}k$ over the types.}
\label{tab:measures}
\centering
\footnotesize
\setlength{\tabcolsep}{3pt}
\begin{tabular}{@{}>{\raggedright\arraybackslash}p{0.15\linewidth}>{\raggedright\arraybackslash}p{0.08\linewidth}>{\raggedright\arraybackslash}p{0.13\linewidth}>{\raggedright\arraybackslash}p{0.15\linewidth}>{\raggedright\arraybackslash}p{0.13\linewidth}>{\raggedright\arraybackslash}p{0.12\linewidth}>{\raggedright\arraybackslash}p{0.12\linewidth}@{}}
\toprule
Measure & Bound to the budget & Credits breadth over types & Adjusts for type size & Caps over-represen\-tation & Order within the budget & Risk weights \\
\midrule
$\mathrm{P@}k$ & yes & no & no & no & no & no \\
Intent-aware precision & yes & no (orders as $\mathrm{P@}k$) & no & no & no & yes \\
Macro recall at $k$ & yes & yes & inversely (favours small types) & only at $n_t$ & no & possible \\
$\mathrm{TC@}k$ & yes & yes (binary) & not needed & at one hit & no & possible \\
$\alpha$-nDCG at $k$ & yes & yes & no & soft decay & yes & in intent-aware variant \\
$\mathrm{FSR@}k$ & yes & yes & yes (fair shares) & at the fair share & no (left to $\mathrm{FHR}_t$) & yes \\
\bottomrule
\end{tabular}
\end{table}

\subsection{Aggregation over seeds and uncertainty}\label{sec:eval-aggregation}

Benchmark results are averaged over seeds, and the type-aware measures do not all average in the same way, which matters for how the results in Section~\ref{sec:results} must be read. $\mathrm{P@}k$ and $R_t\mathrm{@}k$ are linear in the counts, so their seed mean can be computed from seed-averaged counts without loss. $\mathrm{FSR@}k$ is a weighted sum of minima of linear functions and hence concave in the counts; by Jensen's inequality, the FSR of seed-averaged counts is therefore at least the mean of the per-seed FSR values, with equality whenever no type exceeds its cap in any seed, which always holds when every cap equals $n_t$. Type coverage computed from seed-averaged counts a type as covered when any seed covers it and therefore overstates the typical coverage. Two estimands must therefore be kept apart. The FSR of the seed-averaged type profile, $\mathrm{FSR}(\mathbb{E}[f])$, describes the type composition that a detector delivers on average over repeated runs, whereas the expected FSR of a single run, $\mathbb{E}[\mathrm{FSR}(f)]$, describes what one run delivers; the two coincide when every cap equals $n_t$, and otherwise the first is an upper bound on the second. On the public ledger, where the full rankings are available, we report $\mathbb{E}[\mathrm{FSR}(f)]$, computed per seed and then averaged. On the client ledgers only seed-averaged per-type recalls were retained, so every client-ledger FSR in this paper is the FSR of the seed-averaged type profile: it equals the expected single-run FSR on ledgers C and A, where the caps equal the type sizes, and bounds it from above on ledgers B and D, and the intervals reported for B and D reflect the rounding of the retained recalls, not the gap between the two estimands. Coverage computed from seed means would overstate typical coverage and is not reported for the client ledgers (Section~\ref{sec:bench-protocol}). Uncertainty over the units is assessed on the public ledger by resampling its entries with replacement and recomputing every measure on the same resamples for every detector, so that differences between detectors are paired (\ref{app:bootstrap}); consistency across settings is summarised by mean ranks, with the Friedman test and the Nemenyi critical difference \citep{demsar2006statistical} used descriptively because the settings share ledgers.

\subsection{Reading a type profile}\label{sec:eval-reading}

A type profile is read by asking which types reach the reviewer, how close each comes to its ceiling and how early it appears. A per-type recall near its ceiling marks a saturated type, and a zero marks a blind spot at this budget, whatever the hit count. Where every cap equals $k/T$, a large gap between $\mathrm{P@}k$ and $\mathrm{FSR@}k$ shows that the hits pile up on few types, and ledger D illustrates the point: each of its five types holds more anomalies than a budget of 100 can take, the fair share is 20 per type, and a list whose 98 hits are almost all of one type earns about one fifth of the attainable score (Section~\ref{sec:res-typeaware}). The measures are complements rather than substitutes, and we do not propose a single type-aware score. $\mathrm{P@}k$, $\mathrm{FSR@}k$, $\mathrm{TC@}k$ and the first-hit ranks measure different operational properties of a review list, namely its yield, its balance across types within the budget, its breadth and its speed of exposure, and together they form an evaluation profile. FSR can agree with $\mathrm{P@}k$ when the hits are already spread across types, as on the public ledger, where two detectors with low precision nevertheless reach every marking long before the leader in hits (Section~\ref{sec:res-budget}); coverage and first-hit ranks then carry the difference. Conversely, coverage alone would credit a list that touches every type once and misses most anomalies. Equal weights treat every type as an equal risk. Where an engagement's risk assessment ranks the types, for example because a fraud risk points to postings on unusual accounts, the audit team can raise their weights, so that the evaluation rewards the lists that serve the assessed risks; the benchmark ledgers come without such an assessment, which is why we report equal weights throughout. Because the measures need only scores and typed labels, they can also serve as a model-selection criterion wherever a labelled pilot engagement or a benchmark ledger is available, although this use is not tested here.

\section{Benchmark}\label{sec:benchmark}

The benchmark has to meet two requirements that pull in different directions. The type of every anomaly must be known, which calls for controlled injection, and the data must have the texture of real accounting records, which calls for real ledgers whose genuine anomalies are unlabelled. We meet both by injecting typed synthetic anomalies into real client ledgers and by adding a public synthetic ledger whose anomalies were defined by its own generator. On both sources we compare three learning regimes under one protocol that was frozen before the experiments.

\subsection{Client ledgers and typed injection}\label{sec:bench-client}

The four client ledgers are general ledger exports provided by DATEV eG, one fiscal year of one client each and anonymised at source. Ordered by size they are ledgers C, A, B and D, with 3,461 to 279,975 postings (Table~\ref{tab:data}). Besides an identifier, each posting has ten feature fields: account, contra account, currency, tax code and user as anonymised codes, the payment period in days, the modified z-score of the amount and three binary flags (text present, document date outside the fiscal year, round amount). The exports carry no timestamps, no posting texts and no link from a payment to the invoice it clears, so every detector sees one posting at a time. They also contain many exact duplicates: 3.9 to 19.8 percent of the postings repeat all ten feature values of an earlier posting, and up to 912 postings per ledger sit in groups of six or more identical rows, in which the distance to the fifth nearest neighbour is zero. Genuine anomalies that nobody has labelled carry label 0, so every precision on these ledgers is a lower bound and every recall refers to the injected anomalies only.

\begin{table}[tbp]
\caption{Profile of the four client ledgers and the public synthetic ledger. Counts of accounts, contra accounts, tax codes and users are numbers of distinct values. A duplicate row repeats all feature values of another row. Anomalous units are the injected anomalies of the core grid (client ledgers) and the generator-labelled journal entries (public ledger).}
\label{tab:data}
\centering
\footnotesize
\setlength{\tabcolsep}{4pt}
\begin{tabular}{@{}lrrrrr@{}}
\toprule
 & \multicolumn{4}{c}{Client ledgers} & Public \\
\cmidrule(lr){2-5}
Property & C & A & B & D & ledger \\
\midrule
Scored unit & posting & posting & posting & posting & line \\
Evaluated unit & posting & posting & posting & posting & entry \\
Scored rows & 3,461 & 5,080 & 17,505 & 279,975 & 256,232 \\
Evaluated units & 3,461 & 5,080 & 17,505 & 279,975 & 102,429 \\
Fiscal years & 1 & 1 & 1 & 1 & 2 \\
\addlinespace
Accounts & 18 & 23 & 48 & 24 & 25 \\
Contra accounts & 159 & 608 & 362 & 12,359 & 21$^{a}$ \\
Tax codes & 2 & 3 & 2 & 3 & 2 \\
Users & 10 & 2 & 5 & 6 & 6 \\
Payment period missing (\%) & 8.1 & 8.9 & 5.3 & 8.1 & 0.0$^{b}$ \\
\addlinespace
\multicolumn{6}{@{}l}{\textit{Duplicate rows}} \\
\quad repeating an earlier row (\%) & 6.1 & 19.8 & 12.1 & 3.9 & 0.5 \\
\quad in a duplicate group (\%) & 10.1 & 32.6 & 18.5 & 6.9 & 0.6 \\
\quad in groups of six or more & 40 & 51 & 633 & 912 & 1,062 \\
\addlinespace
Anomalous units & 52 & 76 & 263 & 4,200 & 850 \\
Anomalous units (\%) & 1.50 & 1.50 & 1.50 & 1.50 & 0.83 \\
Anomaly classes & T1 to T5 & T1 to T5 & T1 to T5 & T1 to T5 & M1 to M8 \\
\bottomrule
\end{tabular}
\par\smallskip\parbox{\linewidth}{\scriptsize Percentages of duplicate rows refer to the scored rows. $^{a}$ Derived: the account of the largest-amount line on the opposite side of the same entry. $^{b}$ Entry lag, the counterpart of the payment period, is never missing.}
\end{table}

Five anomaly types are injected (Table~\ref{tab:types}): an extreme amount (T1), an extreme payment period (T2), an unseen contra account (T3), an unseen account (T4) and an unusual account pair (T5). In the typology of \citet{foorthuis2021nature}, T1 and T2 are extreme tail values, T3 and T4 unusual classes and T5 an uncommon combination of classes, so the five types address different inductive biases. Each injection modifies one sampled posting in place, which keeps all other fields of the posting real, in line with the argument for realistic rather than uniformly sampled outliers \citep{steinbuss2021benchmarking}, and no posting carries more than one type. The contamination $c$, the share of modified postings, is 0.5, 1.5 or 3 percent, split over the types in fixed shares. Three difficulty levels set the injected amount and payment period to a decreasing multiple of the ledger's 99.9th percentile and lower the frequency that each account of an unusual pair must have. T3 and T4 have a single level: they replace the contra account or the account by a fresh identifier that occurs nowhere in the ledger, one per posting, while T5 writes an account pair that never occurs although each of its accounts does. The core grid runs all four ledgers at $c=1.5$ percent and medium difficulty with five seeds, and the sweep crosses the three contamination rates with the three difficulty levels on ledgers C, A and B, nine cells with five seeds each.

\begin{table}[tbp]
\caption{Anomaly types and markings. (a) Types injected into the client ledgers, with their subtype in the typology of \citet{foorthuis2021nature}, the difficulty parameter for the levels easy, medium and hard, and the number $n_t$ of injected rows in the core grid. (b) Markings of the public ledger, the rule population into which each was injected, and $n_t$.}
\label{tab:types}
\centering
\footnotesize
\setlength{\tabcolsep}{4pt}
\begin{tabular}{@{}>{\raggedright\arraybackslash}p{0.22\linewidth}>{\raggedright\arraybackslash}p{0.47\linewidth}r@{}}
\toprule
\multicolumn{3}{@{}l}{\textit{(a) Types injected into the client ledgers}} \\
Type (Foorthuis subtype) & Modification of a sampled posting & $n_t$: C / A / B / D \\
\midrule
T1 extreme amount (Ia) & amount z-score set to $\pm f\,P_{99.9}(|z|)$ with a random sign; $f = 5, 2, 1$ & 22 / 31 / 105 / 1,680 \\
T2 extreme payment period (Ia) & payment period set to $f\,P_{99.9}$, rows with an observed period only; $f = 5, 2, 1$ & 10 / 15 / 53 / 840 \\
T3 unseen contra account (IIa) & contra account replaced by a fresh id, one distinct id per row & 4 / 6 / 21 / 336 \\
T4 unseen account (IIa) & account replaced by a fresh id, one distinct id per row & 4 / 6 / 21 / 336 \\
T5 unusual account pair (Va) & (account, contra account) replaced by a pair that never occurs in the ledger and whose members each occur at least $F$ times; $F = 50, 20, 5$ & 12 / 18 / 63 / 1,008 \\
\bottomrule
\end{tabular}

\medskip
\begin{tabular}{@{}>{\raggedright\arraybackslash}p{0.25\linewidth}>{\raggedright\arraybackslash}p{0.56\linewidth}lr@{}}
\toprule
\multicolumn{4}{@{}l}{\textit{(b) Markings of the public ledger, defined by its generator}} \\
Marking & Definition & Rule & $n_t$ \\
\midrule
M1 high amount & one of the 1,000 highest-amount entries of a fiscal year, account and user changed into an unusual combination & J1 & 120 \\
M2 high cash disbursement & cash posting above EUR 1,500 with the user changed & J2 & 60 \\
M3 delayed entry & entry lag of 10 to 29 days (less prompt class), usually with an unusual user & J3 & 120 \\
M4 very delayed entry & entry lag above 29 days (not at all prompt class), usually with an unusual user & J3 & 120 \\
M5 non-working hours & posting outside working hours in an under-represented user pattern & J4 & 100 \\
M6 weekend & weekend posting with an unusual user and account context & J4 & 250 \\
M7 shadow bank account & posting rerouted to a rarely used second bank account & none & 60 \\
M8 cross-linked clearing & clearing posting whose reference points to a business transaction of an unrelated category & none & 20 \\
\bottomrule
\end{tabular}
\par\smallskip\parbox{\linewidth}{\scriptsize Subtypes: Ia extreme tail value, IIa unusual class, Va uncommon class combination. Rule populations: J1 top-n amounts, J2 high cash, J3 delayed entry, J4 weekend or non-working hours.}
\end{table}

\subsection{Public synthetic ledger}\label{sec:bench-public}

Injected anomalies carry the risk that a benchmark rewards detectors for finding what its designers put in; the public ledger removes this risk for one source and makes the results checkable by anyone. The DIB General Ledger Data \citep{gronewald2026generating,gronewald2026dib} is a labelled synthetic ledger of a fictional trading company for gardening tools over two fiscal years, with 256,232 posting lines in 102,429 journal entries (Table~\ref{tab:data}). Its generator labels 850 entries, 0.83 percent, with one of eight markings (Table~\ref{tab:types}): high amount (M1), high cash disbursement (M2), delayed entry (M3), very delayed entry (M4), non-working hours (M5), weekend (M6), shadow bank account (M7) and cross-linked clearing (M8). Markings M1 to M6 were injected into the populations that four journal entry testing rules return (J1 top-$n$ amounts, J2 high cash, J3 delayed entry, J4 weekend or non-working hours), whereas M7 and M8 were injected without conditioning on any rule. Each feature field of the client ledgers is mapped to a field of this ledger, with the contra account taken from the largest-amount line on the opposite side of the entry and the payment period replaced by the entry lag, and because this ledger records timestamps, the sine and cosine of the hour and of the weekday are added (Supplementary Table S3).

Detectors score posting lines, whereas the auditor reviews journal entries. An entry $e$ therefore takes the maximum score and the maximum label of its lines,
\begin{equation}
S_e=\max_{i\in e}s_i,\qquad Y_e=\max_{i\in e}y_i,
\label{eq:entry}
\end{equation}
so that an entry is flagged as strongly as its most suspicious line, which is how an auditor would treat it. Labels and markings are constant within an entry, entries are ranked by $S_e$ with the same tie rule as postings, and every measure on this ledger is computed over entries, at the fixed budget of 100 and at $k=1{,}000$, about 1 percent of the entries, as the size-scaled budget of Eq.~\eqref{eq:budget}. The precomputed rule attributes of the dataset enter no feature set: because six markings were injected into rule populations, these columns would partly restate the labels, and the dataset authors recommend the same exclusion. One caveat concerns relations between entries: in the generator a normal payment keeps the user of the invoice it clears, whereas segregation of duties in practice often expects different people to book and to pay, so a change of user is a property of the generator rather than an audit rule.

\subsection{Unsupervised detectors and the supervised reference}\label{sec:bench-detectors}

The unsupervised regime comprises nine detectors chosen to span the main inductive biases of the field (Table~\ref{tab:detectors}). Eight come from PyOD \citep{zhao2019pyod}: isolation forest \citep[IF;][]{liu2008isolation}, which isolates points by random splits; the local outlier factor \citep[LOF;][]{breunig2000lof}, which compares local densities; the distance to the fifth nearest neighbour \citep[kNN;][]{ramaswamy2000efficient}; the histogram-based HBOS \citep{goldstein2012histogram} and the empirical-distribution detector ECOD \citep{li2023ecod}, which score per-field marginal rarity; a principal-component score \citep[PCA;][]{shyu2003novel}; an autoencoder \citep[AE;][]{hinton2006reducing}; and a variational autoencoder \citep[VAE;][]{kingma2014auto}. The ninth, AE-emb, is an autoencoder that learns entity embeddings for the categorical fields \citep{guo2016entity}.

\begin{table}[tbp]
\caption{Detectors, the supervised reference and DeepSAD with their settings, fixed in advance and never tuned on the labels; higher scores mean more anomalous. IF to VAE are PyOD implementations \citep{zhao2019pyod}, AE-emb and DeepSAD are our PyTorch code. $E_{\mathrm{base}}$ is the one-hot and standardised encoding, $E_{\mathrm{emb}}$ the integer-coded input of the embedding autoencoder.}
\label{tab:detectors}
\centering
\footnotesize
\begin{tabular}{@{}>{\raggedright\arraybackslash}p{0.12\linewidth}>{\raggedright\arraybackslash}p{0.2\linewidth}>{\raggedright\arraybackslash}p{0.1\linewidth}>{\raggedright\arraybackslash}p{0.48\linewidth}@{}}
\toprule
Detector & Family & Input & Key settings \\
\midrule
IF & isolation, tree ensemble & $E_{\mathrm{base}}$ & 200 trees, 256 samples per tree \\
LOF & local density & $E_{\mathrm{base}}$ & 20 neighbours, the row itself excluded \\
kNN & distance to neighbours & $E_{\mathrm{base}}$ & distance to the fifth nearest other row \\
HBOS & per-field histogram & $E_{\mathrm{base}}$ & 10 bins, $\alpha = 0.1$, tolerance 0.5 \\
ECOD & empirical tail probability & $E_{\mathrm{base}}$ & no free parameters; computed column-chunked, scores identical to the library \\
PCA & linear projection & $E_{\mathrm{base}}$ & components explaining 90 percent of the variance; score = sum of the distances to the retained eigenvectors divided by their explained-variance ratios \\
AE & reconstruction, neural & $E_{\mathrm{base}}$ & layers 64, 32, 8, 32, 64; ReLU, batch normalisation, dropout 0.2; Adam $10^{-3}$, weight decay $10^{-5}$; 100 epochs, batch 64 \\
VAE & reconstruction, variational & $E_{\mathrm{base}}$ & encoder 64, 32; latent dimension 8; decoder 32, 64; $\beta = 1$, dropout 0.2; 100 epochs, batch 64 \\
AE-emb & reconstruction with entity embeddings & $E_{\mathrm{emb}}$ & embeddings 10, 10, 1, 1, 1; encoder 64, 32, 8 with LeakyReLU 0.01, symmetric decoder; one squared-error head for the numeric block and one cross-entropy head per categorical field; Adam $10^{-3}$; 100 epochs, batch 64 \\
\midrule
LightGBM reference & supervised gradient boosting & $E_{\mathrm{base}}$, labels & library defaults, positive rows weighted by the ratio of negative to positive rows; five-fold stratified cross-fitting, out-of-fold probability as score \\
\midrule
DeepSAD (review protocol) & semi-supervised deep one-class & $E_{\mathrm{emb}}$, reviewed labels & embedding layers and encoder of the AE-emb model of the same setting, initialised with its weights; $\eta = 1$, $\varepsilon = 10^{-6}$; Adam $10^{-4}$, weight decay $10^{-6}$; 50 epochs, batch 128; retrained before every round; round 1 = AE-emb ranking \\
\bottomrule
\end{tabular}
\end{table}

The representation matters as much as the model. The PyOD detectors read the encoding $E_{\mathrm{base}}$, in which amount and payment period are standardised (a missing period imputed and flagged), currency, tax code, user and account are one-hot over all their values, and the contra account enters as its log frequency plus a one-hot of its twenty most frequent values. AE-emb reads $E_{\mathrm{emb}}$, which replaces the one-hot vectors by learned embeddings of integer codes. Two consequences recur in the results: every unseen account (T4) becomes a one-hot column used by a single posting, whereas an unseen contra account (T3) gets no column of its own and is marked only by its frequency of one. All settings are fixed in advance, because an auditor has no labels with which to select hyperparameters, and on ledger D and the public ledger the three networks train for 30 epochs with batch 256 instead of 100 epochs with batch 64 to keep their runtime manageable.

The supervised regime serves as a reference rather than a competitor. A LightGBM model \citep{ke2017lightgbm}, gradient-boosted trees on $E_{\mathrm{base}}$ that read the labels, is trained with five-fold cross-fitting so that each unit is scored by a model fitted on the other folds (on the public ledger with folds grouped by journal entry). Tree ensembles remain strong on tabular data \citep{grinsztajn2022why,shwartzziv2022tabular}, so the failure of this learner on a type indicates that the row-level representation carries little separable signal for it, although it does not prove that the type cannot be learned; the gap between this supervised reference and the unsupervised detectors shows what labels could add.

\subsection{Review protocol with auditor feedback}\label{sec:bench-feedback}

The third regime lets the auditor's reviews feed back into the ranking. In each of four rounds a simulated auditor reviews the 25 highest-ranked units that are still unreviewed; an oracle reveals the labels of the reviewed units only, and the method re-ranks the remaining units with all labels revealed so far, so that after four rounds the auditor has reviewed 100 units, the budget of the static rankings. The method is DeepSAD \citep{ruff2020deep}: unlabelled and reviewed normal units are pulled towards a centre in latent space, reviewed anomalies are pushed away from it, and units are scored by their distance to the centre. DeepSAD appears only in this protocol and not among the static detectors because it needs labels, which the static setting withholds by design: without labels its objective reduces to deep one-class classification, a different unsupervised detector that the frozen protocol did not include, and labels drawn at random in advance would not reflect how an auditor obtains them. Its network consists of the embedding layers and the encoder of the AE-emb model of the same setting and seed, initialised with its trained weights, and before every round after the first it is retrained from these weights with all labels revealed so far (\ref{app:deepsad}). Round 1 therefore reviews the first 25 units of AE-emb, and the first 100 units of AE-emb are the list that the same 100 reviews would have followed had the ranking stayed static. The comparison with this list measures the gain of the adaptive DeepSAD review protocol over the static AE-emb ranking, and that gain bundles two changes: after round 1 the units are scored by their distance to the centre of a retrained latent space instead of by reconstruction error, and the retraining uses the revealed labels. To separate the two, we added in revision a no-feedback control outside the frozen protocol, run on the public ledger with the same initialisation, rounds, seeds and retraining but with the revealed labels withheld, so that every unit enters the retraining unlabelled and the objective takes the label-free form described above. The protocol runs on the core grid of the client ledgers with five seeds and on the public ledger with three, where the oracle reveals entry labels and every line of a reviewed entry carries the label of its entry.

\subsection{Protocol and reproducibility}\label{sec:bench-protocol}

Everything that could favour a detector was fixed before the runs. The design is transductive: each detector is fitted on the injected ledger and scores all of its units, which mirrors how an auditor examines one client-year, and nothing is tuned on the labels. Ties are broken by ascending position in the file, which favours no detector because injected postings are sampled uniformly over positions, and each injected ledger is identified by the SHA-256 hash of a canonical export. On the client ledgers the seeds 0 to 4 drive the injection, the stochastic detectors and the cross-fitting folds; the public ledger is fixed, so its seeds 0 to 2 vary only IF, the three networks and the supervised reference. On the public ledger the type-aware measures are computed per seed from the full rankings and averaged. On the client ledgers the retained results hold per-type recalls at $k=100$ only as seed means printed to two decimals, so the reported $\mathrm{FSR@}100$ is the FSR of the seed-averaged type profile, computed from the seed-averaged counts with the bounds that the rounding implies; by the argument of Section~\ref{sec:eval-aggregation} it equals the expected single-run FSR on ledgers C and A and bounds it from above on ledgers B and D, and type coverage is not reported for these ledgers. All client-ledger results come from a complete run by a second analyst; compared cell by cell with an earlier partial run on another machine, IF, LOF, kNN, HBOS, ECOD, PCA and the supervised reference agree in every compared cell, while the three neural detectors deviate by up to 0.09 in the core-grid means and by up to 0.10 in a per-type recall (Supplementary Section S3). Supplementary Table S2 lists the protocol deviations and Supplementary Table S10 the runtimes.

\section{Results}\label{sec:results}

We report the results in the order of the research questions: the conventional, hit-based view (RQ1), the type-aware view on the client ledgers (RQ2), the effect of the budget on the public ledger (RQ3 and RQ4), the effect of contamination and difficulty (RQ3) and learning from the auditor's reviews (RQ5), followed by robustness checks. Unless stated otherwise, client-ledger results refer to the core grid (contamination 1.5 percent, medium difficulty) as means over five seeds, and public-ledger results average three seeds for the stochastic detectors.

\subsection{The conventional view}\label{sec:res-conventional}

By hit count the comparison has a clear winner on each ledger, but not the same one (Table~\ref{tab:conventional}). kNN has the highest $\mathrm{P@}100$ on ledgers C, A and B, with 0.40, 0.29 and 0.42, and the highest AP on the same three ledgers; only on ledger C do VAE and AE come within one or two hits. IF, LOF, HBOS and ECOD stay close to chance there, with a $\mathrm{P@}100$ of at most 0.08 against 0.015 for a random ordering and an AP of at most 0.11. Ledger D reverses the order: PCA, AE and VAE place 98 anomalies among their first 100 postings, kNN places 34, and AE has the highest AP. On the public ledger every unsupervised detector struggles; AE-emb has the highest $\mathrm{P@}100$, 0.40, but varies strongly between seeds (Section~\ref{sec:res-robust}), kNN follows at 0.17 with the highest AP, and the other seven stay at or below 0.11. The supervised reference reaches a $\mathrm{P@}100$ of 0.97 or more on ledgers B and D and 0.83 on the public ledger, which indicates that with labels most anomalies there are separable from single rows, whereas on ledger C it does no better than kNN.

\begin{table}[tbp]
\caption{Conventional, hit-based view: P@100 and average precision (AP), mean (standard deviation) over seeds. Client ledgers: core grid, five seeds; supervised reference as a mean only. Public ledger: three seeds for IF, AE, VAE, AE-emb and the supervised reference, one run for the deterministic detectors. Bold: best unsupervised detector per column. The supervised LightGBM reference uses the labels and is shown for comparison only.}
\label{tab:conventional}
\centering
\footnotesize
\setlength{\tabcolsep}{3pt}
\begin{tabular}{@{}lrrrrr@{}}
\toprule
 & \multicolumn{4}{c}{Client ledgers (five seeds)} & Public \\
\cmidrule(lr){2-5}
Detector & C & A & B & D & ledger \\
\midrule
\multicolumn{6}{@{}l}{\textit{(a) P@100}} \\
IF & 0.04 (0.01) & 0.03 (0.02) & 0.04 (0.04) & 0.11 (0.07) & 0.08 (0.04) \\
LOF & 0.06 (0.03) & 0.01 (0.01) & 0.00 (0.00) & 0.11 (0.03) & 0.11 \\
kNN & \textbf{0.40 (0.02)} & \textbf{0.29 (0.04)} & \textbf{0.42 (0.03)} & 0.34 (0.04) & 0.17 \\
HBOS & 0.03 (0.01) & 0.02 (0.02) & 0.03 (0.03) & 0.04 (0.01) & 0.04 \\
ECOD & 0.08 (0.01) & 0.03 (0.01) & 0.08 (0.03) & 0.05 (0.01) & 0.04 \\
PCA & 0.32 (0.02) & 0.15 (0.02) & 0.22 (0.01) & \textbf{0.98 (0.00)} & 0.03 \\
AE & 0.38 (0.01) & 0.22 (0.06) & 0.23 (0.00) & \textbf{0.98 (0.00)} & 0.02 (0.01) \\
VAE & 0.39 (0.02) & 0.16 (0.02) & 0.24 (0.03) & \textbf{0.98 (0.00)} & 0.02 (0.01) \\
AE-emb & 0.12 (0.02) & 0.19 (0.01) & 0.23 (0.07) & 0.27 (0.11) & \textbf{0.40 (0.23)} \\
\addlinespace
\textit{LightGBM reference} & 0.40 & 0.56 & 0.97 & 0.99 & 0.83 (0.09) \\
\midrule
\multicolumn{6}{@{}l}{\textit{(b) AP}} \\
IF & 0.04 (0.01) & 0.03 (0.01) & 0.03 (0.01) & 0.04 (0.00) & 0.05 (0.01) \\
LOF & 0.11 (0.02) & 0.04 (0.01) & 0.02 (0.00) & 0.06 (0.00) & 0.03 \\
kNN & \textbf{0.52 (0.02)} & \textbf{0.25 (0.04)} & \textbf{0.36 (0.02)} & 0.22 (0.00) & \textbf{0.12} \\
HBOS & 0.03 (0.01) & 0.03 (0.00) & 0.02 (0.00) & 0.04 (0.00) & 0.05 \\
ECOD & 0.05 (0.01) & 0.04 (0.00) & 0.05 (0.00) & 0.08 (0.00) & 0.06 \\
PCA & 0.23 (0.02) & 0.09 (0.01) & 0.12 (0.00) & 0.29 (0.00) & 0.05 \\
AE & 0.31 (0.03) & 0.19 (0.02) & 0.18 (0.01) & \textbf{0.35 (0.01)} & 0.06 (0.01) \\
VAE & 0.44 (0.13) & 0.16 (0.01) & 0.19 (0.01) & 0.30 (0.01) & 0.06 (0.00) \\
AE-emb & 0.09 (0.02) & 0.16 (0.03) & 0.14 (0.03) & 0.13 (0.01) & 0.08 (0.02) \\
\addlinespace
\textit{LightGBM reference} & 0.67 & 0.73 & 0.86 & 0.97 & 0.74 (0.02) \\
\bottomrule
\end{tabular}
\par\smallskip\parbox{\linewidth}{\scriptsize Precision on the client ledgers is a lower bound, because their genuine anomalies carry label 0 (Section~\ref{sec:bench-client}).}
\end{table}

To summarise how consistently the detectors rank across conditions, we compute their mean ranks by $\mathrm{P@}100$ over the 27 cells of the contamination and difficulty sweep on ledgers C, A and B (Section~\ref{sec:res-sweep}), each entered as its mean over five seeds because the per-seed values were not retained, and apply the Friedman test with the Nemenyi post-hoc comparison \citep{demsar2006statistical}. Because the 27 conditions derive from only three ledgers, they are not independent datasets in the sense the test assumes, so the test and the critical-difference diagram are exploratory: we read them as a description of rank consistency rather than as inference about a population of ledgers. The mean ranks differ strongly ($\chi^2=195.9$), kNN ranks first with a mean rank of 1.15, VAE (2.72) and AE (3.00) lie within the Nemenyi critical difference of 2.31 of it, and AE-emb, PCA, ECOD, IF, LOF and HBOS lie outside it (Fig.~\ref{fig:sweep}b). By hits, then, the distance-based detector is the most consistent performer on the three smaller ledgers, the two neural reconstruction models are close behind, and the isolation, density and per-field detectors fail. The type-aware view asks what these hits consist of.

\subsection{The type-aware view on the client ledgers}\label{sec:res-typeaware}

\begin{figure}[tbp]
\centering
\includegraphics[width=\textwidth]{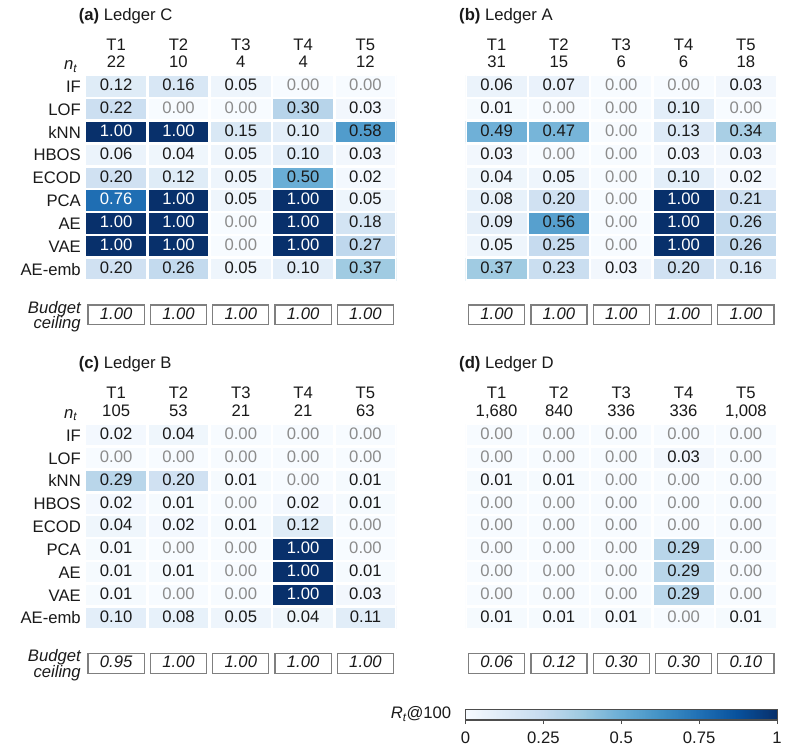}
\caption{Recall within the first 100 postings per injected type, $R_t$@100, on the four client ledgers (core grid, mean over five seeds): (a) ledger C, (b) ledger A, (c) ledger B, (d) ledger D. Column headers give the type and its number $n_t$ of injected rows. The framed bottom row is the budget ceiling $\min(1, 100/n_t)$.}
\label{fig:pertype}
\end{figure}

The per-type recalls in Fig.~\ref{fig:pertype} show that each detector family finds the anomaly types that match its inductive bias. The budget ceiling in the bottom row of the figure is 1 or close to 1 on ledgers C, A and B, where the budget can hold every anomaly of a type, and at most 0.30 on ledger D. kNN, which scores distance to the nearest neighbours, finds the extreme values: on ledger C it recovers every extreme amount (T1) and every extreme payment period (T2), on ledger A about half of them, and on both ledgers it finds more unusual account pairs (T5) than any other detector. PCA, AE and VAE, which score reconstruction error on the one-hot encoding, find the unseen accounts (T4): they recover all of them on ledgers C, A and B, and on ledger D their T4 recall of 0.29 nearly reaches its ceiling of 0.30. On ledger C, AE and VAE also recover all T1 and T2 rows, but on ledger B they find almost nothing else, with a recall of at most 0.03 on the other four types. AE-emb, which embeds the account codes instead of one-hot coding them, spreads its hits thinly, with a recall above zero for every type on C and A but never above 0.37, and IF, LOF, HBOS and ECOD find few anomalies of any type; their highest recall, 0.5 for ECOD on ledger C, amounts to two of the four unseen accounts there. The unseen contra account (T3) escapes every family: the best T3 recall is 0.15 on C, 0.03 on A, 0.05 on B and 0.01 on D, although on the first three ledgers the budget could hold every T3 row.

\begin{figure}[tbp]
\centering
\includegraphics[width=\textwidth]{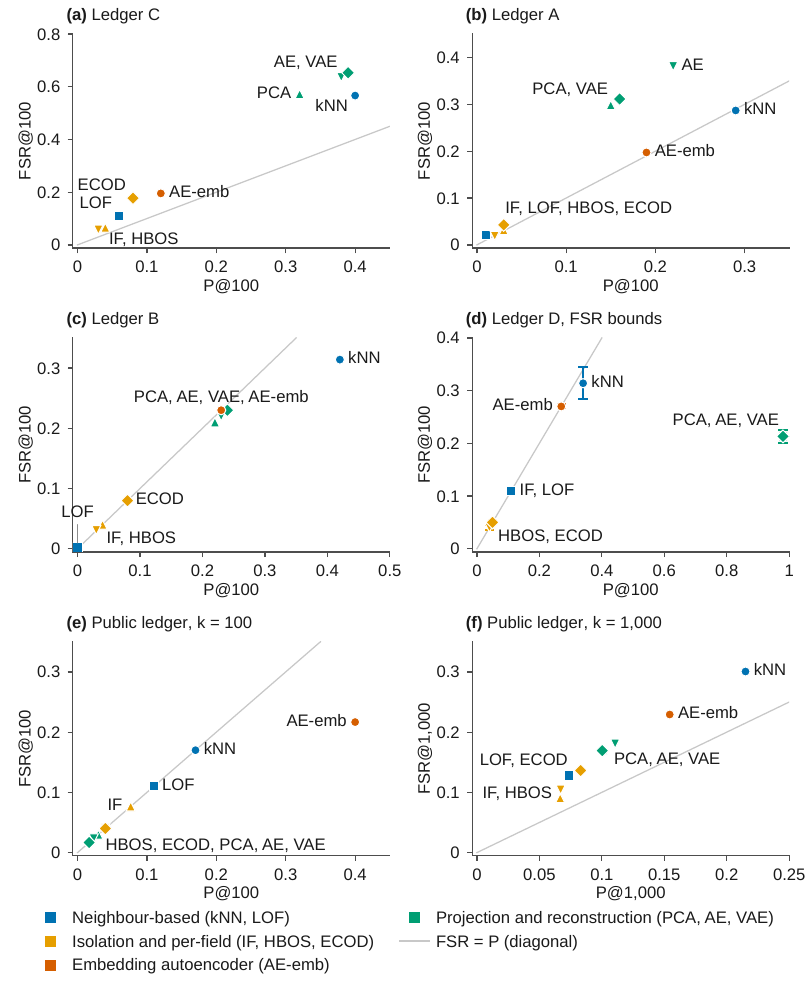}
\caption{Hits against fair-share type recall per detector: P@$k$ on the horizontal and FSR@$k$ on the vertical axis; the grey diagonal marks FSR = P. (a) to (d) Ledgers C, A, B and D at $k=100$ (seed means; on ledger D, bars show the rounding bounds). (e), (f) The public ledger at $k=100$ and $k=1{,}000$. Axis ranges differ between panels.}
\label{fig:hitstypes}
\end{figure}

Fair-share type recall turns these profiles into a comparison. On the client ledgers it is computed for the seed-averaged type profile (Section~\ref{sec:eval-aggregation}). On ledgers C and A the budget exceeds the number of anomalies, so $\mathrm{FSR@}100$ is the mean per-type recall; on all client ledgers it is known only within the bounds that the two-decimal rounding implies, to about 0.01 on C, A and B and less precisely on D, and type coverage is not reported (Section~\ref{sec:bench-protocol}). Fig.~\ref{fig:hitstypes} plots $\mathrm{FSR@}100$ against $\mathrm{P@}100$, and Table~\ref{tab:typeaware} summarises what changes: on three of the four ledgers the detector with the most hits is not the one with the highest FSR, and each of these reversals holds for every value inside the rounding bounds. On ledger C, kNN has the most hits ($\mathrm{P@}100=0.40$) but an $\mathrm{FSR@}100$ of 0.57, whereas VAE, with one hit fewer, reaches 0.65, and AE and, narrowly, PCA also rank above kNN; the unseen accounts decide it, since VAE, AE and PCA find all four and kNN hardly any. On ledger A, kNN again has the most hits and AE the highest FSR, 0.38 against 0.29, with VAE and PCA also ahead of kNN and AE-emb, third by hits, behind PCA and VAE. On ledger B, kNN leads both views with an $\mathrm{FSR@}100$ between 0.31 and 0.32, and Kendall's $\tau_b$ between the orderings by $\mathrm{P@}100$ and by $\mathrm{FSR@}100$ is 0.97 or higher.

\begin{table}[tbp]
\caption{What changes under the type-aware view. Per ledger: the detector with the highest P@$k$, its type coverage TC@$k$ and its FSR@$k$; the detector with the highest FSR@$k$; and Kendall's $\tau_b$ between the orderings of the nine detectors by P@$k$ and by FSR@$k$. Client ledgers: FSR of the seed-averaged type profile, with ranges given by the bounds that the two-decimal rounding implies; public ledger: mean of the per-seed values.}
\label{tab:typeaware}
\centering
\footnotesize
\setlength{\tabcolsep}{3pt}
\begin{tabular}{@{}lllllllr@{}}
\toprule
 & \multicolumn{4}{c}{Best by hits} & \multicolumn{2}{c}{Best by FSR@$k$} & \\
\cmidrule(lr){2-5}\cmidrule(lr){6-7}
Ledger ($k$) & Detector & P@$k$ & TC@$k$ & FSR@$k$ & Detector & FSR@$k$ & $\tau_b$ \\
\midrule
C (100) & kNN & 0.40 & n/d & 0.57 & VAE & 0.65 & 0.83 \\
A (100) & kNN & 0.29 & n/d & 0.29 & AE & 0.38 & 0.65 \\
B (100) & kNN & 0.42 & n/d & 0.31--0.32 & kNN & 0.31--0.32 & 0.97--0.99 \\
D (100) & \makecell[l]{PCA, AE,\\VAE} & 0.98 & n/d & 0.200--0.226 & kNN & 0.284--0.344 & 0.598--0.625 \\
Public (100) & AE-emb & 0.40 & 2.7 & 0.22 & AE-emb & 0.22 & 1.00 \\
Public (1,000) & kNN & 0.215 & 6 & 0.301 & kNN & 0.301 & 1.00 \\
\bottomrule
\end{tabular}
\par\smallskip\parbox{\linewidth}{\scriptsize n/d: not determined by the seed-mean recalls of the client ledgers (Section~\ref{sec:bench-protocol}). TC@$k$ counts types out of eight markings. Public ledger: exact, from the full rankings, mean over three seeds for the stochastic detectors.}
\end{table}

Ledger D shows the largest change, with $\tau_b$ between 0.60 and 0.63. Every type there holds at least 336 anomalies, so no per-type recall can exceed 0.30 within 100 postings (Fig.~\ref{fig:pertype}d). PCA, AE and VAE fill this budget with a single type: at least 95.8 of their 98 hits are unseen accounts, and because hits beyond the fair share of 20 earn nothing, their $\mathrm{FSR@}100$ is at most 0.226. kNN finds about a third as many anomalies but spreads them over at least two types and reaches between 0.284 and 0.344, and AE-emb, whose hits cover at least four types, also ranks above the three hit leaders. Apart from ties this ordering is fixed by the bounds (Supplementary Tables S6 and S7). It concerns the seed-averaged type profiles; for the expected FSR of a single run the bound of 0.226 on the three hit leaders still holds, by Jensen's inequality, but kNN's single-run value cannot be bounded from below with the retained results. The answer to RQ2 is therefore that the type-aware view changes which detector ranks highest on three of four client ledgers, in both directions: towards the reconstruction models on the two smaller ledgers, where they add the unseen accounts that kNN misses, and towards kNN on the largest, where the reconstruction models find almost nothing but unseen accounts.

Which of these conclusions require the fair-share normalisation, and which follow from reading hits by type at all? Table~\ref{tab:macro} compares the leaders and orderings under $\mathrm{P@}100$, macro-averaged recall, the plain mean of $R_t\mathrm{@}100$ over the types, and $\mathrm{FSR@}100$, all computed from the same retained results. On ledgers C and A the budget holds every anomaly, every cap equals $n_t$ and FSR coincides with macro recall ($\tau_b=1.00$): the reversals there demonstrate the value of a type-aware reading in general, and any average over types would reveal them. On ledgers B and D the caps are active and the two measures part. On ledger D, macro recall keeps PCA, AE and VAE on top, at 0.058 against 0.004 for kNN, because it values a hit of type $t$ at $1/(T n_t)$ and so rewards the 336 unseen accounts, the smallest types there, most; only the fair-share caps turn the concentration of their hits into a limit and place kNN first ($\tau_b=0.42$ between macro recall and FSR). On ledger B, conversely, macro recall would promote VAE and AE over kNN, at about 0.21 against 0.10, on the strength of the 21 unseen accounts they find, whereas FSR credits these only up to the fair share of 20 and keeps kNN in the lead. The reversal on the largest ledger thus rests on the fair-share normalisation, and on ledger B the same normalisation prevents macro recall from rewarding concentration on the smallest type; on the public ledger the three measures agree closely.

\begin{table}[tbp]
\caption{Which conclusions need the fair-share normalisation: the leader under P@$k$, macro-averaged recall (the mean of $R_t$@$k$ over the types) and FSR@$k$, the fair-share caps ($n_t$ when every cap equals the type size, in which case FSR equals macro recall, otherwise the common cap $k/T$), and Kendall's $\tau_b$ between the orderings of the nine detectors. Client ledgers at $k=100$, from the printed seed-mean per-type recalls (FSR of the seed-averaged type profile, midpoints of the rounding bounds; detectors within 0.005 of the macro-recall leader are listed together). Public ledger exact, mean over seeds.}
\label{tab:macro}
\centering
\footnotesize
\setlength{\tabcolsep}{2.8pt}
\begin{tabular}{@{}lllllcc@{}}
\toprule
 & & \multicolumn{3}{c}{Leader by} & \multicolumn{2}{c}{Kendall's $\tau_b$} \\
\cmidrule(lr){3-5}\cmidrule(l){6-7}
Ledger ($k$) & Cap $c_t$ & P@$k$ & Macro & FSR@$k$ & P, macro & macro, FSR \\
\midrule
C (100) & $n_t$ & kNN & VAE & VAE & 0.83 & 1.00 \\
A (100) & $n_t$ & kNN & AE & AE & 0.65 & 1.00 \\
B (100) & 20 & kNN & VAE, AE & kNN & 0.74 & 0.69 \\
D (100) & 20 & PCA, AE, VAE & PCA, AE, VAE & kNN & 0.81 & 0.42 \\
Public (100) & 12.5 & AE-emb & AE-emb & AE-emb & 0.94 & 0.94 \\
Public (1,000) & $n_t$ & kNN & kNN & kNN & 1.00 & 1.00 \\
\bottomrule
\end{tabular}
\end{table}

\subsection{The budget on the public ledger}\label{sec:res-budget}

\begin{figure}[tbp]
\centering
\includegraphics[width=\textwidth]{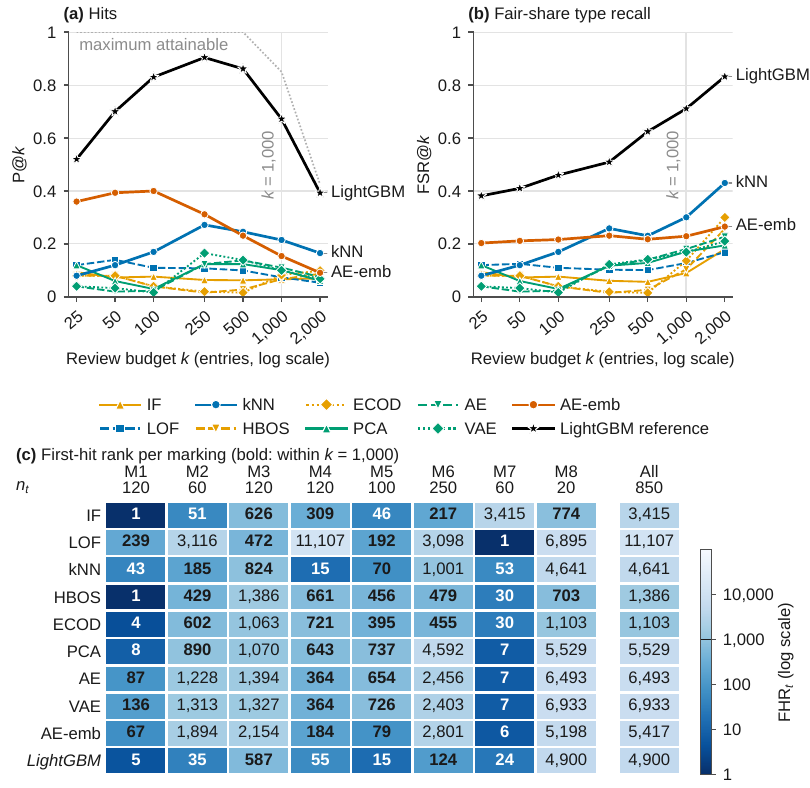}
\caption{The review budget on the public ledger (102,429 journal entries, 850 anomalous). (a) P@$k$ and (b) FSR@$k$ against the budget $k$ (log scale); the dotted line in (a) is the attainable maximum and the vertical line marks $k=1{,}000$. (c) First-hit rank per marking (dark: early; bold: within 1,000); All is the rank at which every marking has appeared.}
\label{fig:budget}
\end{figure}

Because the full rankings of the public ledger are available, the type-aware view can be traced across budgets (Fig.~\ref{fig:budget}a): AE-emb has the most hits up to $k=250$ and kNN from $k=500$ on, reaching $\mathrm{P@}1{,}000=0.215$ at the size-scaled budget. $\mathrm{FSR@}k$ tells the same story (Fig.~\ref{fig:budget}b and Fig.~\ref{fig:hitstypes}e,f): at $k=100$ and $k=1{,}000$ it orders the nine detectors exactly as $\mathrm{P@}k$ does ($\tau_b=1.0$), and it names a different leader only at $k=250$, kNN instead of AE-emb. FSR penalises concentration here as well, but not enough to change any pairwise order. Type coverage and first-hit ranks, however, tell a different story. Kendall's $\tau_b$ between $\mathrm{P@}k$ and $\mathrm{TC@}k$ is 0.50 at $k=100$ and $-0.17$ at $k=1{,}000$, so at the larger budget more hits do not mean more markings: HBOS has among the fewest hits within 1,000 entries, with $\mathrm{P@}1{,}000$ of 0.067, yet covers seven of the eight markings, one more than kNN.

First-hit ranks show the same effect from the auditor's side (Fig.~\ref{fig:budget}c). HBOS and ECOD show all eight markings within 1,386 and 1,103 entries, although their first 250 entries show only two, whereas kNN shows seven markings within 1,001 entries but its first cross-linked clearing (M8) only at rank 4,641; the supervised reference shows seven markings within 587 entries on average and no M8 entry within its first 1,000 (Supplementary Table S4). What fills the top of the rankings explains why the per-field and reconstruction detectors start so slowly. For HBOS, ECOD, PCA, AE and VAE, 96 to 99 of the first 100 entries (seed 0) are legitimate entries that touch one of 17 rare accounts, those with fewer than 1,000 lines; only 577 normal entries touch these accounts, yet they fill almost the entire top 100 of these five detectors, whereas the first 100 entries of kNN hold 17 anomalies and six such entries (Supplementary Table S8). Rarity-based models, in other words, find what is rare, and on this ledger what is rarest is mostly legitimate.

\begin{figure}[tbp]
\centering
\includegraphics[width=\textwidth]{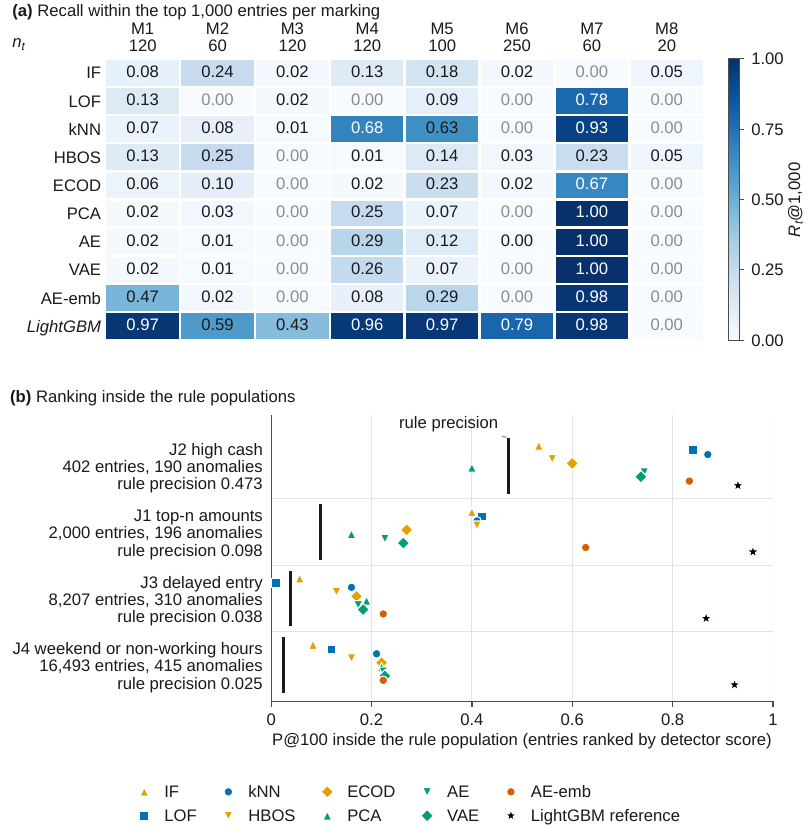}
\caption{Markings and rule populations on the public ledger. (a) Recall per marking within the first 1,000 journal entries; $n_t$ is the number of anomalous entries per marking. (b) P@100 when the entries inside each journal entry testing rule population are ranked by the detector score; the black bar marks the rule precision, the share of anomalies in the population.}
\label{fig:publictypes}
\end{figure}

The recall per marking within the first 1,000 entries (Fig.~\ref{fig:publictypes}a) completes the picture. PCA, AE and VAE find every shadow bank account entry (M7), a posting rerouted to a rarely used bank account; kNN finds most very delayed entries (M4), postings outside working hours (M5) and M7 entries; and AE-emb finds about half of the high amounts (M1). Three markings stay out of reach: for every unsupervised detector, recall within the first 1,000 entries is at most 0.05 for delayed entries (M3), weekend postings (M6) and cross-linked clearings (M8). The supervised reference finds M3 and M6 (recall 0.43 and 0.79) but no M8 entry, so even with labels a row-level model does not separate cross-linked clearings. A post-hoc diagnostic, run outside the frozen protocol and not part of the benchmark results, locates the missing signal: scoring each entry that clears an earlier entry by the number of fields (user, item text, gross amount) in which it differs from the referenced original places all 20 M8 entries and 92 percent of the M6 entries within its first 1,000. Inside the four journal entry testing rule populations, finally, the best unsupervised detector ranks entries well above plain rule matching (Fig.~\ref{fig:publictypes}b); in the weekend or non-working-hours population, for example, 0.23 of its 100 highest-scored entries are anomalous, about nine times the rule precision of 0.025 (Supplementary Table S9).

\subsection{Contamination and difficulty}\label{sec:res-sweep}

\begin{figure}[tbp]
\centering
\includegraphics[width=\textwidth]{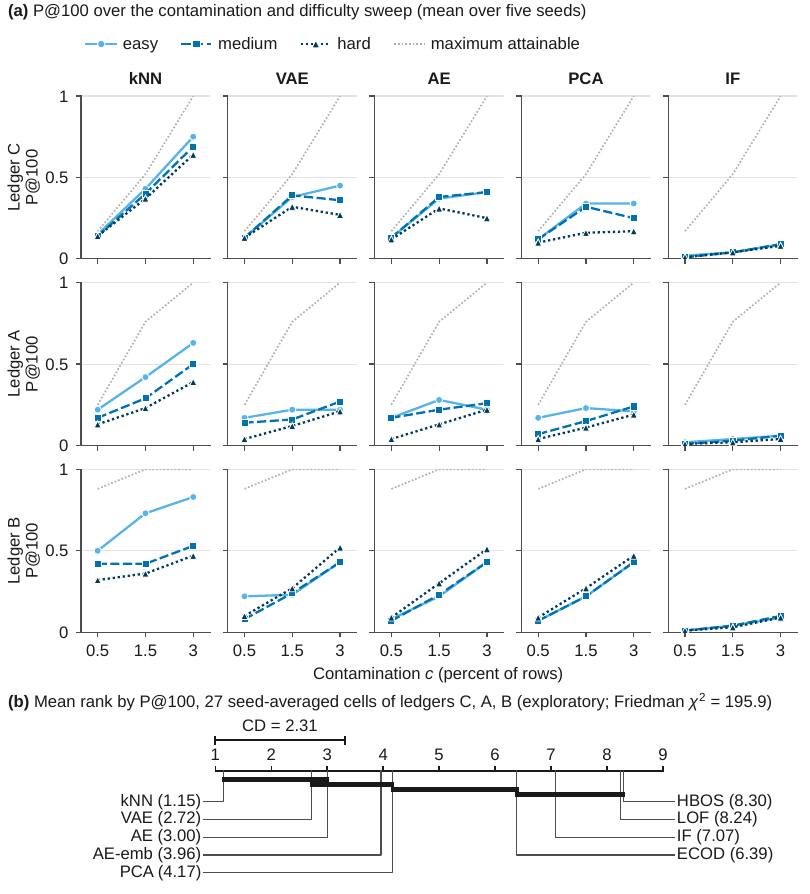}
\caption{Contamination and difficulty on ledgers C, A and B. (a) P@100 of kNN, VAE, AE, PCA and IF at contamination 0.5, 1.5 and 3 percent and three difficulty levels (mean over five seeds); the dotted line is the attainable maximum. (b) Exploratory critical-difference diagram of mean ranks over the 27 seed-averaged sweep cells, which derive from three ledgers and are not independent datasets; bars join detectors within CD = 2.31.}
\label{fig:sweep}
\end{figure}

The sweep asks whether the hit-based comparison depends on how many anomalies a ledger contains and how pronounced they are (Fig.~\ref{fig:sweep}a). Part of the rise of $\mathrm{P@}100$ with contamination is mechanical, because at 0.5 percent every ledger holds fewer than 100 anomalies and the attainable maximum caps the measure. kNN and IF gain with contamination on all three ledgers, and PCA, AE and VAE roughly double their $\mathrm{P@}100$ on ledger B from 1.5 to 3 percent, but on C and A they gain at most 0.11 over the same step and in some cells lose. Difficulty separates the families more clearly. kNN loses in every setting as the anomalies become harder, by 0.11 on ledger C, 0.24 on A and 0.36 on B from the easy to the hard level at 3 percent contamination, as expected when milder extreme values sit closer to their neighbours. PCA, AE and VAE hold or gain on ledger B at 1.5 and 3 percent as the anomalies get harder, since T4, the type they find best, has no difficulty level, whereas on C at 1.5 and 3 percent and on A at 1.5 percent they lose. IF stays at or below 0.10 in every cell. Across all cells kNN has the highest $\mathrm{P@}100$ in 24 of 27 and ties in one more; the exceptions are ledger A at 1.5 percent and hard difficulty, where AE-emb leads by one hit, and ledger B at 3 percent and hard difficulty, where VAE leads (Supplementary Table S11). Because per-type recalls were retained for the core grid only, the sweep speaks to hits; how contamination and difficulty move the type profiles remains untested.

\subsection{Auditor reviews and the type gaps}\label{sec:res-feedback}

The review protocol asks whether labels from the auditor's own reviews improve the list, and whether the improvement widens the range of types or only deepens the types already found. Table~\ref{tab:feedback}a counts the anomalies found in 100 reviews. Round 1 reviews the first 25 units of AE-emb and so finds what AE-emb finds; from round 2 on, the network is retrained with the labels revealed so far. After 100 reviews DeepSAD has found 29.4, 47.0, 57.6 and 76.0 anomalies on ledgers C, A, B and D and 68.3 on the public ledger, whereas the first 100 units of AE-emb, the list that the same reviews would have followed had the ranking stayed static, hold about 12, 19, 23, 27 and 40. The adaptive review protocol thus adds about 17 to 49 anomalies per 100 reviews over the static AE-emb ranking on every ledger, and on the public ledger it also removes most of the seed dependence of the starting model: the three AE-emb seeds hold 19, 37 and 64 anomalies among their first 100 entries, and DeepSAD ends at 71, 64 and 70. Against the static detector with the most hits the picture is mixed. The protocol wins on ledgers A and B, by 18.0 and 15.6 anomalies over kNN, loses on ledgers C and D, by 10.6 to kNN and 22.0 to PCA, AE and VAE, and on the public ledger exceeds AE-emb, the static leader at this budget, by 28.3 while staying below the 83 of the supervised reference. Because the static leader is identified with the labels, this comparison shows what the protocol has to beat rather than what an auditor could choose in advance.

\begin{table}[tbp]
\caption{Auditor reviews fed back to DeepSAD. (a) Anomalies found after 25, 50, 75 and 100 reviews (four rounds of 25), mean over seeds with the standard deviation of the final count, against the anomalies among the first 100 units of AE-emb, the list the same reviews follow if the ranking stays static, and of the static detector with the most hits; the gain is that of the adaptive review protocol over the static AE-emb ranking. Client ledgers: core grid, five seeds; the static counts are 100 times the printed P@100 and exact to within 0.5. Public ledger: three seeds, units are journal entries. Public, control: the same protocol with the revealed labels withheld from the retraining, added in revision outside the frozen protocol. (b) Public ledger: markings among the 100 reviewed entries, with and without labels, and among the first 100 entries of AE-emb, mean over three seeds, with type coverage TC@100 and fair-share type recall FSR@100; $n_t$ is the number of entries per marking.}
\label{tab:feedback}
\centering
\footnotesize
\setlength{\tabcolsep}{3.4pt}
\begin{tabular}{@{}lrrrrrrll@{}}
\multicolumn{9}{@{}l}{\textit{(a) Anomalies found in 100 reviews}} \\
\toprule
 & \multicolumn{4}{c}{DeepSAD, after $r$ reviews} & \multicolumn{2}{c}{AE-emb, first 100} & \multicolumn{2}{c}{Most hits, first 100} \\
\cmidrule(lr){2-5}\cmidrule(lr){6-7}\cmidrule(l){8-9}
Ledger & 25 & 50 & 75 & 100 & Found & Gain & Detector & Found \\
\midrule
C & 3.0 & 21.0 & 26.2 & 29.4 (4.6) & 12 (2) & 17.4 & kNN & 40 \\
A & 8.8 & 29.0 & 40.6 & 47.0 (3.0) & 19 (1) & 28.0 & kNN & 29 \\
B & 8.0 & 23.4 & 39.4 & 57.6 (11.7) & 23 (7) & 34.6 & kNN & 42 \\
D & 10.2 & 29.4 & 52.8 & 76.0 (8.9) & 27 (11) & 49.0 & PCA, AE, VAE & 98 \\
Public & 9.0 & 27.7 & 50.7 & 68.3 (3.8) & 40.0 (22.7) & 28.3 & AE-emb & 40.0 \\
Public, control & 9.0 & 22.0 & 33.7 & 40.3 (18.6) & 40.0 (22.7) & 0.3 & AE-emb & 40.0 \\
\bottomrule
\end{tabular}

\vspace{1.2em}
\setlength{\tabcolsep}{3.2pt}
\begin{tabular}{@{}lrrrrrrrrrrr@{}}
\multicolumn{12}{@{}l}{\textit{(b) Public ledger: markings among 100 entries}} \\
\toprule
 & M1 & M2 & M3 & M4 & M5 & M6 & M7 & M8 & Hits & TC@100 & FSR@100 \\
\midrule
$n_t$ & 120 & 60 & 120 & 120 & 100 & 250 & 60 & 20 & 850 & 8 & \\
AE-emb, first 100 & 24.0 & 0 & 0 & 1.0 & 2.3 & 0 & 12.7 & 0 & 40.0 & 2.7 & 0.22 \\
DeepSAD, reviewed & 24.0 & 0 & 0 & 2.0 & 2.3 & 0 & 40.0 & 0 & 68.3 & 3.0 & 0.23 \\
Control, reviewed & 19.7 & 0 & 0 & 0.7 & 1.7 & 0 & 18.3 & 0 & 40.3 & 3.3 & 0.23 \\
\bottomrule
\end{tabular}
\end{table}

This gain belongs to the adaptive protocol as a whole and not yet to the auditor's labels, because the protocol changes two things at once: after round 1 it scores the units by their distance to a centre in a retrained latent space instead of by reconstruction error, and the retraining uses the revealed labels. The no-feedback control added in revision (Section~\ref{sec:bench-feedback}) keeps the first change and removes the second. On the public ledger it ends at 40.3 anomalies after 100 reviews, level with the 40.0 of the static AE-emb list, whereas DeepSAD with labels reaches 68.3 (Table~\ref{tab:feedback}a), so that on average the gain over the static list is carried by the labels, which add 10, 29 and 45 anomalies over the control in the three seeds. Retraining without labels is nevertheless far from neutral: it moves the count from 19 to 61 in seed 0, from 37 to 35 in seed 1 and from 64 to 25 in seed 2 (Table~\ref{tab:app-deepsad-seeds}), so the unlabelled rescoring gains or loses up to about 40 anomalies depending on the seed, and it is the labels that stabilise the outcome at 64 to 71. On the client ledgers the control was not run, and the gains there remain those of the adaptive protocol over the static AE-emb ranking, which may contain a retraining component of either sign. With three seeds on one ledger, we read the control as evidence that the labels matter, not as an exact decomposition of the gain.

The type-aware view qualifies the gain. On the public ledger the marking of every reviewed entry is known, and Table~\ref{tab:feedback}b compares the 100 reviewed entries with the first 100 entries of AE-emb and with the entries reviewed in the control: the hits rise from 40.0 to 68.3, but type coverage rises only from 2.7 to 3.0 markings and $\mathrm{FSR@}100$ from 0.22 to 0.23, and the control, with 3.3 markings and an $\mathrm{FSR@}100$ of 0.23, reaches at least as many markings with 28 fewer hits. In every seed one marking takes most of the reviewed hits (Table~\ref{tab:app-deepsad-seeds}). Shadow bank accounts (M7) account for 60 of 71 hits in seed 0 and 55 of 70 in seed 2, against 18 and 15 among the first 100 entries of AE-emb, and high amounts (M1) account for 58 of 64 hits in seed 1, against 26; in seed 2 the M1 entries even fall from 46 to 9 as M7 takes their place, and $\mathrm{FSR@}100$ rises in seed 0 but falls in seeds 1 and 2. The labels sharpen this concentration: without them the most frequent marking takes 35 of 61, 28 of 35 and 15 of 25 reviewed hits, or 57 to 80 percent, and with them 79 to 91 percent. No reviewed entry in any seed, with or without labels, is a high cash disbursement (M2), a delayed entry (M3), a weekend posting (M6) or a cross-linked clearing (M8). The answer to RQ5 is thus that the adaptive review protocol raises the number of anomalies found and that, on the public ledger, the auditor's labels carry this rise on average, but they do not widen the range of markings reached; we read this as evidence of an exploitation bias in ranking-derived feedback, not as a general limit of semi-supervised learning. For the client ledgers only the number of anomalies found per round was retained, not their types, so whether the reviews concentrate there as well remains open.

\subsection{Robustness}\label{sec:res-robust}

Three checks bear on how far the comparisons can be trusted: resampling of the public ledger, the stability of the stochastic detectors across seeds and the dependence of FSR on the type weights. We resampled the 102,429 public-ledger entries with replacement 1,000 times, used the same resamples for every detector and took the seed-0 scores of the stochastic detectors (\ref{app:bootstrap}). kNN is ahead of every other unsupervised detector in all 1,000 resamples, on $\mathrm{P@}1{,}000$ and on $\mathrm{FSR@}1{,}000$ alike, whereas at $k=100$ the 95 percent percentile interval of the paired difference contains zero against AE-emb and LOF and excludes it against the other six detectors. AE-emb is the least stable detector across seeds: on the public ledger its three seeds give $\mathrm{P@}100$ of 0.19, 0.37 and 0.64, and on ledgers B and D it has the largest standard deviation of $\mathrm{P@}100$ over five seeds (Table~\ref{tab:conventional}). kNN's lead at the size-scaled budget is therefore robust to resampling, whereas its comparison with AE-emb at a budget of 100 is not resolved.

Because equal weights are a convention, we also recomputed FSR under alternative weights from the retained results (\ref{app:weights}). When the weight of one type is doubled or halved, the reversal on ledger A persists in all eleven schemes, equal weights included, and on ledger C in all but the one that halves the weight of unseen accounts, the type that separates VAE from kNN there; on ledger D it remains fixed by the rounding bounds in nine schemes, and on ledger B a reversal towards VAE appears in two. Across 300 random weight vectors with every weight between half and twice its equal value, the reversal holds in 91, 89 and 63 percent of the vectors on ledgers C, A and D and appears in 26 percent on B, while on the public ledger kNN stays the FSR leader at $k=1{,}000$ in 98 percent of 2,000 vectors. The reversals are therefore robust to moderate departures from equal weights but not to every weighting: a risk assessment that discounts the type separating two models can legitimately change which of them is preferred.

\section{Discussion}\label{sec:discussion}

A hit-based and a type-aware comparison of the same rankings can recommend different models, and the difference follows from how inductive bias, representation and the source of the labels interact. We summarise what the type-aware view changes, explain the mechanisms behind it, draw implications for the design of AI-based journal entry testing and state the limits of the evidence.

\subsection{What the type-aware view changes}\label{sec:disc-changes}

By hits (RQ1), kNN is the strongest unsupervised detector on the three smaller client ledgers and at the size-scaled budget of the public ledger. PCA, AE and VAE lead on ledger D, and the isolation, density and per-field detectors stay close to chance. By types (RQ2), the top-ranked detector changes on three of the four client ledgers, in both directions; on ledgers C and A any average over types reveals the change, whereas on ledger D only the fair-share normalisation does, and the unseen contra account escapes every family. The budget moves the type-aware view most (RQ3): at $k=1{,}000$ the orderings by hits and by coverage on the public ledger are unrelated, whereas contamination and difficulty shift hit counts without changing the leader in most settings. The public ledger (RQ4) reproduces the ease of accounts used only by anomalies for PCA and the autoencoders and kNN's lead at a size-scaled budget, and it adds rare legitimate entries at the top of the rankings and a relational blind spot. The adaptive review protocol (RQ5) raises the number of anomalies found over its static starting ranking, on the public ledger on average through the auditor's labels, but where types can be traced it barely widens the range of markings. The choice of a model for journal entry testing therefore depends on which question the evaluation asks, and a hit-based benchmark answers only one of them.

\subsection{Learning mechanisms behind the results}\label{sec:disc-mechanisms}

The first mechanism is the gap between rarity and relevance. Histogram, empirical-distribution and reconstruction models rank by statistical rarity, and in a ledger most rare entries are legitimate: on the public ledger at least 96 of the first 100 entries of HBOS, ECOD, PCA, AE and VAE are normal entries on rarely used accounts such as wages, rent and utilities (Supplementary Table S8), which echoes the observation that statistically unusual records need not be the interesting ones \citep{pelleg2004active}. Even the shadow bank account (M7), which only anomalies use, barely enters these lists, although PCA, AE and VAE find all its entries within 1,000. On the client ledgers we measured how often postings repeat (Table~\ref{tab:data}) but not what fills the top ranks, so whether rarity also explains the near-chance precision of IF, LOF, HBOS and ECOD there remains untested; distance concentration in high-dimensional encodings \citep{beyer1999nearest,zimek2012survey}, which can make a rare legitimate posting as isolated as an injected anomaly, is a further unmeasured candidate.

The second mechanism is representation. Each unseen account (T4) becomes a one-hot column used by a single posting, the rarest value in the ledger, and PCA, AE and VAE find every unseen account within 100 postings on ledgers C, A and B, whereas AE-emb, which embeds accounts instead, reaches a recall of at most 0.20 on this type. An unseen contra account (T3) gets no column of its own, because the contra account enters as its log frequency plus a one-hot of its most frequent values, and it is found by no detector. The same model thus sees or misses a type depending on how the field is encoded, which supports ADBench's conclusion that performance depends on the match between a detector's assumptions and the anomaly \citep{han2022adbench} and adds that the representation is part of that match.

The third mechanism is the limit of row-level learning. On the public ledger delayed (M3), weekend (M6) and cross-linked (M8) entries reach a recall of at most 0.05 within the first 1,000 entries for every unsupervised detector (Fig.~\ref{fig:publictypes}). Their evidence lies in the document chain: depending on the marking, 74 to 100 percent of the anomalous clearing entries are booked by a different user than the invoice they clear, against 1 percent of the normal ones, and every M8 entry clears an original with a different amount. The supervised reference indicates that this is a limit of the representation rather than of the model: it also scores one row at a time and misses every M8 entry, whereas the post-hoc relational diagnostic places them all within its first 1,000. The user change is a generator convention (Section~\ref{sec:bench-public}); what transfers is where the signal lies, in the relation between entries.

The fourth mechanism concerns learning from the expert. Labels taken from the top of a ranking are not a representative sample of the data \citep{dasgupta2008hierarchical}; they concentrate on the points that the ranking already puts first \citep{pelleg2004active}, and a model retrained on them moves more of the same kind upwards. The public-ledger results fit this account, and the no-feedback control sharpens it: the labels add on average 28 anomalies over the same protocol without labels, but with them 55 to 60 of the 64 to 71 reviewed hits fall on a single marking that the starting list already held, a larger share than without them, and the reviewed lists reach no more markings with labels than without (3.0 against 3.3 on average). Feedback in this form exploits what the model already knows about the anomalies rather than exploring what it has not yet seen; with three seeds on one ledger, we read this as evidence of an exploitation bias in ranking-derived feedback rather than as a general property of semi-supervised learning.

\subsection{Implications for AI-based journal entry testing}\label{sec:disc-implications}

The first implication concerns evaluation: a model for expert decision support should be judged by which labelled anomaly categories its review list reaches, and hence by which risks where the categories correspond to them, not only by its hits, and the judgement should rest on the profile of complementary measures rather than on any single score. A hit rate of 0.98 on ledger D would pass any hit-based acceptance test, yet the list tests one risk; the type-aware measures expose this at no extra cost, because they need only the scores and the types of labelled anomalies that any benchmark or pilot engagement already has. Second, the budget should scale with ledger size, and first-hit ranks should inform how far an audit team reads. Third, the complementary type profiles, with kNN strong on extreme values and pairs, the reconstruction models on unseen accounts and the per-field detectors quick to reach every marking on the public ledger, suggest that models should be chosen, and perhaps combined, by the risks they cover; we did not test combinations, and the measures proposed here are the natural criterion for doing so. Fourth, representation deserves as much attention as model choice: relational risks require features that link entries, such as the invoice a payment clears, its users and the amount difference, and until detectors have them, rule-based tests on document chains should stay in the test plan. Fifth, systems that learn from the auditor's reviews should report their type profile alongside their hits and should be designed to explore types that no reviewed entry represents, for instance through type-aware query strategies; and where an engagement's risk assessment ranks the types, as ISA 330 asks the auditor to respond to assessed risks \citep{iaasb2025isa330}, the FSR weights, and possibly the selected model, should follow it; the weight analysis in Section~\ref{sec:res-robust} shows that such a choice can matter when it down-weights a type that separates the models.

\subsection{Limitations}\label{sec:disc-limitations}

The evidence has limits of four kinds. Construct validity is limited by the labels: they are injected on the client ledgers and generator-defined on the public one, genuine client anomalies carry label 0, so that precision there is a lower bound, and injected types may be easier or harder to find than real misstatements; equal type weights, moreover, are a risk-neutral convention that stands in for a risk assessment the benchmark ledgers lack, and the weight analysis bounds but does not remove this dependence; the types themselves are anomaly categories, which correspond to audit risks only where an engagement defines them so. Internal validity is limited by the retained results. On the client ledgers the type-aware measures rest on rounded seed means, so the reported FSR is that of the seed-averaged type profile: it equals the expected single-run FSR on ledgers C and A but only bounds it from above on ledgers B and D, where the per-seed type counts that would identify the single-run value were not retained, and type coverage is not reported. The neural detectors, moreover, are not reproducible across machines at the printed precision: AE, VAE and AE-emb agree in fewer than half of the compared cells of two independent executions and move by up to 0.09 in the core grid (Supplementary Section S3); the reversals on ledgers C and A do not hinge on them, because kNN also falls below the deterministic PCA by FSR there, but results that rest on a single neural detector should be read with this variability in mind. External validity is limited by the data: four clients with one fiscal year each and structured fields only, and a public ledger that is synthetic and comes from the first author's group, so that it is not an independent replication; the client ledgers cannot be shared, which confines independent checks to the public ledger. Statistical conclusion validity, finally, is limited by the Friedman test on seed-averaged cells that derive from three ledgers and are therefore not independent datasets, by bootstrap intervals that exist for the public ledger only and by a type concentration under feedback that is shown on the public ledger with three seeds, because the types of the reviewed rows on the client ledgers were not retained; the no-feedback control that separates the contribution of the labels from that of the retraining was added in revision and exists for the public ledger only, so that on the client ledgers the gains are those of the adaptive protocol as a whole. Future work can address these limits with feedback strategies that seek types not yet seen, with document-chain features for relational risks and with evaluations weighted by the risk assessments of real engagements.

\section{Conclusion}\label{sec:conclusion}

Anomaly detectors for journal entry testing are used as expert systems in which a model ranks the entries and an auditor reviews the first $k$, yet they are evaluated as if every anomaly were the same. This paper introduced a type-aware budget evaluation that asks which kinds of anomaly reach the reviewer, applied it to four real client ledgers and a public synthetic ledger, and compared unsupervised, supervised and expert-feedback learning. The type-aware view changes which model ranks highest on three of the four client ledgers for their seed-averaged type profiles. On the two smaller ledgers any average over types reveals this; on the largest, only the fair-share normalisation does, placing three models that fill the budget with hits of one type below a detector with a third of their hits, where macro-averaged recall would keep them on top. On the public ledger, detectors with few hits reach all anomaly types sooner than the hit leader, rarity-based models fill their top ranks with legitimate rare entries, one relational anomaly type escapes row-level learning even with labels, and the auditor's labels in an adaptive review loop mostly multiply the hits of the types already found rather than reaching new ones. These findings follow from the inductive biases of the models, from the representation of the data and from the way expert labels are sampled, and they are invisible to hit-based measures. Evaluations of machine learning for audit should therefore report which labelled anomaly categories, and so which risks where the categories map to them, a model's review list reaches, as a profile of complementary measures rather than a single score and not only as a count of anomalies, and the design of such systems should treat representation and the exploration of unseen types as first-class concerns.

\section*{CRediT authorship contribution statement}

\textbf{Jan Gronewald:} Conceptualization, Methodology, Software, Data curation, Resources, Visualization, Writing - original draft, Writing - review \& editing, Project administration, Funding acquisition. \textbf{Michel Scherer:} Software, Validation, Investigation. \textbf{Nijat Mehdiyev:} Conceptualization, Methodology, Formal analysis, Writing - original draft, Writing - review \& editing.

\section*{Declaration of competing interest}

Jan Gronewald is a co-author of the synthetic ledger dataset used in this study. The authors declare no other known competing financial interests or personal relationships that could have appeared to influence the work reported in this paper.

\section*{Data availability}

The four client ledgers were provided by DATEV eG in anonymised form under a research agreement and cannot be shared publicly. The synthetic ledger is publicly available on Zenodo (\url{https://doi.org/10.5281/zenodo.22014653}) under a CC BY 4.0 licence. The code that reproduces all results on the synthetic ledger, and the aggregated results on the client ledgers, are available from the corresponding author on reasonable request.

\section*{Declaration of generative AI and AI-assisted technologies in the writing process}

During the preparation of this work the authors used Claude (Anthropic) in order to assist with code development, analysis scripting and drafting of the manuscript text. After using this tool, the authors reviewed and edited the content as needed and take full responsibility for the content of the publication.

\section*{Acknowledgements}

We thank DATEV eG for providing the anonymised general ledger data. This research was funded by the German Federal Ministry of Research, Technology and Space (BMFTR) within the Software Campus programme (grant number 01IS23064).

\appendix
\makeatletter\@addtoreset{table}{section}\@addtoreset{figure}{section}\makeatother
\setcounter{table}{0}\setcounter{figure}{0}

\section{Notation}\label{app:notation}

Table~\ref{tab:app-notation} collects the symbols used in Sections~\ref{sec:evaluation} and \ref{sec:benchmark} and in \ref{app:deepsad}. A unit is a posting on the client ledgers and a journal entry on the public ledger, and every measure is computed on the ranking that a detector's scores induce, with ties broken by ascending position in the file. The symbols that refer to types apply equally to the injected types T1 to T5 of the client ledgers and to the generator-defined markings M1 to M8 of the public ledger, both of which the evaluation treats as labelled anomaly categories (Section~\ref{sec:eval-hits}); they measure audit risks only where an engagement maps its risks to these categories.

\begin{table}[tbp]
\caption{Notation of the type-aware budget evaluation, the benchmark and the review protocol.}
\label{tab:app-notation}
\centering
\footnotesize
\begin{tabular}{@{}l>{\raggedright\arraybackslash}p{0.74\textwidth}@{}}
\toprule
Symbol & Meaning \\
\midrule
$N$, $y_i$ & number of units; label of unit $i$ (1 anomalous, 0 normal) \\
$\mathcal{T}$, $T$ & set of anomaly types (T1 to T5, or markings M1 to M8) and its size \\
$\mathcal{I}_t$, $n_t$ & anomalies of type $t$ and their number \\
$s_i$, $\pi(r)$ & score of unit $i$; unit at rank $r$ (descending score, ties by position) \\
$k$, $\rho$ & review budget; budget as a share of the ledger, $k=\rho N$ \\
$\operatorname{top}_k$, $f_t(k)$ & the first $k$ units of the ranking; number of anomalies of type $t$ among them \\
$\mathrm{P@}k$, $R_t\mathrm{@}k$ & precision at $k$; recall of type $t$ at $k$, bounded by the budget ceiling $\min(1,k/n_t)$ \\
$\mathrm{TC@}k$ & type coverage, the number of types with at least one anomaly among the first $k$ units \\
$c_t$, $s^{*}$, $w_t$ & fair-share cap of type $t$, water level of the caps, weight of type $t$ \\
$\mathrm{FSR@}k$ & fair-share type recall, $\sum_t w_t\min(f_t(k),c_t)/c_t$ \\
$\mathrm{FHR}_t$ & first-hit rank of type $t$; $\max_t\mathrm{FHR}_t$ is the full-coverage budget \\
$S_e$, $Y_e$ & score and label of journal entry $e$, maxima over its lines \\
$E_{\mathrm{base}}$, $E_{\mathrm{emb}}$ & one-hot and standardised encoding; integer-coded input of the embedding autoencoder \\
$\varphi$, $\boldsymbol{\omega}$, $\tilde{y}_i$ & DeepSAD network, centre, review label ($-1$ reviewed anomaly, $+1$ reviewed normal unit, 0 unreviewed) \\
$\eta$, $\varepsilon$ & weight of the labelled term and stabilising constant of the DeepSAD loss \\
\bottomrule
\end{tabular}
\end{table}

\section{DeepSAD in the review protocol}\label{app:deepsad}

The review protocol of Section~\ref{sec:bench-feedback} uses DeepSAD \citep{ruff2020deep} with a network $\varphi$ that consists of the embedding layers and the encoder, including bias terms, of the AE-emb model of the same setting and seed, initialised with its trained weights $\varphi_0$. This choice ties the feedback model to its unsupervised starting point, so that the first 100 units of AE-emb are the list that the same reviews would have followed had the ranking stayed static. With $\mathbf{z}_i$ the $E_{\mathrm{emb}}$ input of unit $i$, the centre is the mean initial code, with small coordinates pushed away from zero to avoid a trivial solution,
\begin{equation}
\boldsymbol{\omega}=\frac{1}{N}\sum_{i=1}^{N}\varphi_0(\mathbf{z}_i),\qquad
\omega_j\leftarrow 0.1\,\operatorname{sgn}(\omega_j)\ \ \text{if}\ |\omega_j|<0.1,\quad \operatorname{sgn}(0)=+1 .
\label{eq:app-dsad-centre}
\end{equation}
With $\tilde{y}_i=-1$ for a reviewed anomaly, $+1$ for a reviewed normal unit and $0$ for an unreviewed unit, and $\varphi_i=\varphi(\mathbf{z}_i)$, the loss on a mini-batch $\mathcal{M}$ is
\begin{equation}
\mathcal{L}=\frac{1}{|\mathcal{M}|}\sum_{i\in\mathcal{M}}\Bigl(\mathbb{I}[\tilde{y}_i=0]\,\|\varphi_i-\boldsymbol{\omega}\|_2^2+\mathbb{I}[\tilde{y}_i\neq0]\,\eta\,\bigl(\|\varphi_i-\boldsymbol{\omega}\|_2^2+\varepsilon\bigr)^{\tilde{y}_i}\Bigr),
\label{eq:app-dsad-loss}
\end{equation}
with $\eta=1$ and $\varepsilon=10^{-6}$, and units are scored by $\|\varphi_i-\boldsymbol{\omega}\|_2^2$. Unlabelled and reviewed normal units are thus pulled towards the centre and reviewed anomalies pushed away from it.

Round 1 uses the AE-emb ranking. Before every later round the network is retrained from $\varphi_0$ on all units with the labels revealed so far, with a round-specific seed, by Adam with learning rate $10^{-4}$ and weight decay $10^{-6}$ in place of the weight penalty of the original objective, for 50 epochs with mini-batches of 128 and with only the embedding and encoder parameters updated. On the public ledger DeepSAD trains on posting lines, each line of a reviewed entry carrying the label of its entry, and entries are scored by the maximum over their lines, as in Eq.~\eqref{eq:entry}. The oracle reveals only the labels of the reviewed units, and the harness checks in every round that no other label reaches the method. One run of the four rounds on the public ledger took 1,150.0\,s on average over the three seeds. The no-feedback control added in revision runs the same four rounds with $\tilde{y}_i=0$ for every unit in Eq.~\eqref{eq:app-dsad-loss}, so that the loss reduces to the deep one-class objective \citep{ruff2018deep} and the retraining sees no label, while the reviewed units are still removed from the ranking; the centre, the seeds, the optimiser and the schedule are unchanged, and one run took 885.6\,s on average. Table~\ref{tab:app-deepsad-seeds} reports the per-seed results on the public ledger that underlie Section~\ref{sec:res-feedback}: in every seed one marking takes most of the reviewed hits, more so with labels than without, the control moves the count in either direction depending on the seed, and no seed reviews an entry of M2, M3, M6 or M8.

\begin{table}[htbp]
\caption{Public ledger, DeepSAD per seed: markings among the 100 reviewed entries with labels and in the no-feedback control (labels withheld, added in revision), and among the first 100 entries of the AE-emb model both runs start from, with the anomalies found after 25 and 100 reviews (F@25, F@100), type coverage TC@100 and fair-share type recall FSR@100. Round 1 reviews the first 25 entries of AE-emb, so the three lists agree after 25 reviews.}
\label{tab:app-deepsad-seeds}
\centering
\scriptsize
\setlength{\tabcolsep}{3pt}
\begin{tabular}{@{}llrrrrrrrrrrrr@{}}
\toprule
Seed & List & M1 & M2 & M3 & M4 & M5 & M6 & M7 & M8 & F@25 & F@100 & TC@100 & FSR@100 \\
\midrule
0 & AE-emb, first 100 & 0 & 0 & 0 & 0 & 1 & 0 & 18 & 0 & 6 & 19 & 2 & 0.135 \\
0 & DeepSAD, reviewed & 5 & 0 & 0 & 0 & 6 & 0 & 60 & 0 & 6 & 71 & 3 & 0.235 \\
0 & Control, reviewed & 25 & 0 & 0 & 0 & 1 & 0 & 35 & 0 & 6 & 61 & 3 & 0.260 \\
\addlinespace
1 & AE-emb, first 100 & 26 & 0 & 0 & 0 & 6 & 0 & 5 & 0 & 15 & 37 & 3 & 0.235 \\
1 & DeepSAD, reviewed & 58 & 0 & 0 & 0 & 1 & 0 & 5 & 0 & 15 & 64 & 3 & 0.185 \\
1 & Control, reviewed & 28 & 0 & 0 & 0 & 2 & 0 & 5 & 0 & 15 & 35 & 3 & 0.195 \\
\addlinespace
2 & AE-emb, first 100 & 46 & 0 & 0 & 3 & 0 & 0 & 15 & 0 & 6 & 64 & 3 & 0.280 \\
2 & DeepSAD, reviewed & 9 & 0 & 0 & 6 & 0 & 0 & 55 & 0 & 6 & 70 & 3 & 0.275 \\
2 & Control, reviewed & 6 & 0 & 0 & 2 & 2 & 0 & 15 & 0 & 6 & 25 & 4 & 0.225 \\
\bottomrule
\end{tabular}
\end{table}

\section{Bootstrap intervals on the public ledger}\label{app:bootstrap}

\begin{figure}[tbp]
\centering
\includegraphics[width=\textwidth]{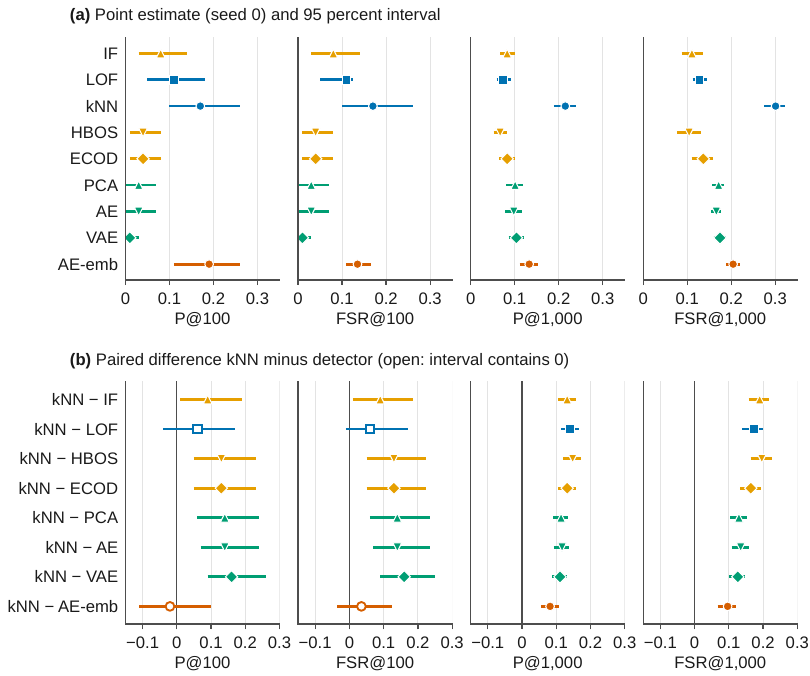}
\caption{Bootstrap intervals on the public ledger: 1,000 resamples of the 102,429 journal entries with replacement, the same resamples for every detector, percentile 95 percent intervals; the stochastic detectors enter with their seed-0 scores. (a) Point estimate and interval of P@100, FSR@100, P@1,000 and FSR@1,000. (b) Paired difference kNN minus each other detector; open markers are intervals that contain zero.}
\label{fig:app-bootstrap}
\end{figure}

The public ledger is a single fixed dataset, so the uncertainty of its results over the units can be assessed by resampling the journal entries. We drew 1,000 resamples of the 102,429 entries with replacement, used the same resamples for every detector, so that differences between detectors are paired, and report percentile 95 percent intervals (Table~\ref{tab:app-bootstrap} and Fig.~\ref{fig:app-bootstrap}); the stochastic detectors enter with their seed-0 scores. At the size-scaled budget of $k=1{,}000$, kNN is ahead of every other unsupervised detector on both $\mathrm{P@}k$ and $\mathrm{FSR@}k$ in all resamples, so its lead there does not depend on which entries the ledger happens to contain. At $k=100$ the paired intervals of kNN against AE-emb and against LOF contain zero, so kNN is not separated from these two detectors at the fixed budget, although it is from the other six. The seed variation of AE-emb, whose $\mathrm{P@}100$ ranges from 0.19 to 0.64 over the three seeds, is wider than any bootstrap interval, so its lead in the three-seed mean reflects the seed as much as the entries. For the client ledgers only seed means and standard deviations were retained (Supplementary Table S1), not the per-seed and per-unit results that a paired bootstrap needs, which is why the resampling analysis is confined to the public ledger.

\begin{table}[tbp]
\caption{Public ledger, bootstrap over journal entries: 1,000 resamples of the entries with replacement, the same resamples for every detector, percentile 95 percent intervals. The stochastic detectors enter with their seed-0 scores, so the point estimates can differ from the three-seed means in Table~\ref{tab:conventional}. (a) Intervals per detector. (b) Paired differences kNN minus the named detector; italic marks an interval that contains zero, and the second line gives the share of resamples in which kNN is strictly ahead.}
\label{tab:app-bootstrap}
\centering
\scriptsize
\setlength{\tabcolsep}{3pt}
\begin{tabular}{@{}lrrrrr@{}}
\toprule
\multicolumn{6}{@{}l}{\textit{(a) Point estimate [95 percent interval]}} \\
\addlinespace
Detector & P@100 & FSR@100 & P@1,000 & FSR@1,000 & TC@1,000 \\
\midrule
IF & 0.08 [0.03, 0.14] & 0.08 [0.03, 0.14] & 0.083 [0.067, 0.100] & 0.111 [0.087, 0.136] & 7 [6, 7] \\
LOF & 0.11 [0.05, 0.18] & 0.11 [0.05, 0.13] & 0.074 [0.059, 0.092] & 0.128 [0.112, 0.145] & 4 [3, 4] \\
kNN & 0.17 [0.10, 0.26] & 0.17 [0.10, 0.26] & 0.215 [0.189, 0.240] & 0.301 [0.275, 0.323] & 6 [5, 7] \\
HBOS & 0.04 [0.01, 0.08] & 0.04 [0.01, 0.08] & 0.067 [0.052, 0.083] & 0.104 [0.077, 0.131] & 7 [5, 7] \\
ECOD & 0.04 [0.01, 0.08] & 0.04 [0.01, 0.08] & 0.083 [0.065, 0.100] & 0.136 [0.110, 0.158] & 6 [5, 6] \\
PCA & 0.03 [0.00, 0.07] & 0.03 [0.00, 0.07] & 0.101 [0.081, 0.119] & 0.171 [0.156, 0.183] & 5 [3, 5] \\
AE & 0.03 [0.00, 0.07] & 0.03 [0.00, 0.07] & 0.098 [0.079, 0.116] & 0.166 [0.154, 0.177] & 4 [3, 5] \\
VAE & 0.01 [0.00, 0.03] & 0.01 [0.00, 0.03] & 0.104 [0.086, 0.122] & 0.174 [0.160, 0.189] & 5 [3, 5] \\
AE-emb & 0.19 [0.11, 0.26] & 0.14 [0.11, 0.17] & 0.133 [0.111, 0.154] & 0.204 [0.189, 0.219] & 4 [4, 4] \\
\bottomrule
\end{tabular}

\bigskip
\begin{tabular}{@{}lrrrr@{}}
\toprule
\multicolumn{5}{@{}l}{\textit{(b) Paired difference kNN minus detector [95 percent interval]; share of resamples with kNN ahead}} \\
\addlinespace
Compared with & P@100 & FSR@100 & P@1,000 & FSR@1,000 \\
\midrule
IF & 0.09 [0.01, 0.19] & 0.09 [0.01, 0.19] & 0.132 [0.105, 0.157] & 0.190 [0.158, 0.217] \\
 & {\scriptsize 97.6\,\%} & {\scriptsize 97.6\,\%} & {\scriptsize 100.0\,\%} & {\scriptsize 100.0\,\%} \\
LOF & \textit{0.06 [$-$0.04, 0.17]} & \textit{0.06 [$-$0.01, 0.17]} & 0.141 [0.114, 0.166] & 0.173 [0.140, 0.200] \\
 & {\scriptsize 87.8\,\%} & {\scriptsize 93.8\,\%} & {\scriptsize 100.0\,\%} & {\scriptsize 100.0\,\%} \\
HBOS & 0.13 [0.05, 0.23] & 0.13 [0.05, 0.23] & 0.148 [0.120, 0.173] & 0.196 [0.164, 0.226] \\
 & {\scriptsize 99.9\,\%} & {\scriptsize 99.9\,\%} & {\scriptsize 100.0\,\%} & {\scriptsize 100.0\,\%} \\
ECOD & 0.13 [0.05, 0.23] & 0.13 [0.05, 0.23] & 0.132 [0.106, 0.158] & 0.164 [0.134, 0.194] \\
 & {\scriptsize 99.9\,\%} & {\scriptsize 99.9\,\%} & {\scriptsize 100.0\,\%} & {\scriptsize 100.0\,\%} \\
PCA & 0.14 [0.06, 0.24] & 0.14 [0.06, 0.24] & 0.114 [0.091, 0.135] & 0.129 [0.104, 0.152] \\
 & {\scriptsize 100.0\,\%} & {\scriptsize 100.0\,\%} & {\scriptsize 100.0\,\%} & {\scriptsize 100.0\,\%} \\
AE & 0.14 [0.07, 0.24] & 0.14 [0.07, 0.24] & 0.117 [0.093, 0.138] & 0.135 [0.109, 0.158] \\
 & {\scriptsize 100.0\,\%} & {\scriptsize 100.0\,\%} & {\scriptsize 100.0\,\%} & {\scriptsize 100.0\,\%} \\
VAE & 0.16 [0.09, 0.26] & 0.16 [0.09, 0.25] & 0.111 [0.089, 0.131] & 0.126 [0.101, 0.148] \\
 & {\scriptsize 100.0\,\%} & {\scriptsize 100.0\,\%} & {\scriptsize 100.0\,\%} & {\scriptsize 100.0\,\%} \\
AE-emb & \textit{$-$0.02 [$-$0.11, 0.10]} & \textit{0.04 [$-$0.04, 0.13]} & 0.082 [0.055, 0.107] & 0.096 [0.067, 0.122] \\
 & {\scriptsize 40.6\,\%} & {\scriptsize 82.1\,\%} & {\scriptsize 100.0\,\%} & {\scriptsize 100.0\,\%} \\
\bottomrule
\end{tabular}
\end{table}

\section{Sensitivity to the type weights}\label{app:weights}

Equal type weights are a risk-neutral benchmarking convention (Section~\ref{sec:eval-fsr}), and a sceptical reader may ask how far the comparisons depend on it. We therefore recomputed FSR under alternative weights from the retained results, without new runs (Table~\ref{tab:app-weights}). On the client ledgers, $\mathrm{FSR@}100$ is bounded exactly as in the main analysis, from the per-type recalls and $\mathrm{P@}100$ printed to two decimals over five seeds, and a reversal counts only when the bounds fix it, that is, when some other detector's lower bound exceeds the upper bound of the $\mathrm{P@}100$ leader. The structured schemes are equal weights and the ten schemes that double or halve the weight of one type, normalised to sum to one, and the random schemes are 300 vectors drawn uniformly from the simplex subject to every weight lying between half and twice its equal value. On the public ledger FSR is exact per seed from the full rankings and averaged over seeds; the structured schemes are equal weights and the sixteen schemes that double or halve one marking, and the random schemes are 2,000 vectors drawn in the same way.

The results separate robust from weight-dependent conclusions. The reversal on ledger A survives every structured scheme, and the reversal on ledger C every scheme except the one that halves the weight of unseen accounts, the type on which VAE and kNN differ most there. On ledger D the reversal remains fixed by the bounds in nine of the eleven schemes, and on ledger B, where equal weights keep kNN in the lead, a reversal towards VAE appears in two. On the public ledger kNN stays the FSR leader at $k=1{,}000$ in all structured schemes and AE-emb at $k=100$ in all but one, and the orderings by $\mathrm{P@}k$ and $\mathrm{FSR@}k$ remain close throughout. A risk assessment that discounts the type separating two models can therefore change which of them is preferred, but moderate departures from equal weights leave the main reversals in place.

\begin{table}[htbp]
\caption{Sensitivity of FSR@$k$ to the type weights (\ref{app:weights}). (a) Client ledgers at $k=100$: the leader by P@100, the FSR leader under equal weights, the number of the eleven structured schemes (equal weights and each type doubled or halved) in which a reversal between the P@100 leader and FSR is fixed by the rounding bounds, the schemes that are exceptions, and the share of 300 random weight vectors (each weight between half and twice its equal value) with a reversal fixed by the bounds. (b) Public ledger: the FSR leader under equal weights, the number of the seventeen structured schemes in which it stays the FSR leader, the smallest Kendall $\tau_b$ between the orderings by P@$k$ and FSR@$k$ over these schemes, and, over 2,000 random vectors, the share in which the equal-weight FSR leader stays the leader and the median $\tau_b$.}
\label{tab:app-weights}
\centering
\scriptsize
\setlength{\tabcolsep}{3pt}
\begin{tabular}{@{}l>{\raggedright\arraybackslash}p{0.13\textwidth}>{\raggedright\arraybackslash}p{0.13\textwidth}>{\raggedright\arraybackslash}p{0.12\textwidth}>{\raggedright\arraybackslash}p{0.26\textwidth}>{\raggedright\arraybackslash}p{0.14\textwidth}@{}}
\multicolumn{6}{@{}l}{\textit{(a) Client ledgers, $k=100$}} \\
\toprule
Ledger & P@100 leader & FSR leader, equal weights & Reversal fixed (of 11 schemes) & Exceptions & Random vectors with reversal fixed \\
\midrule
C & kNN & VAE & 10 & T4 halved & 91\% \\
A & kNN & AE & 11 & none & 89\% \\
B & kNN & kNN (no reversal) & 2 & reversal to VAE when T1 is halved or T4 doubled & 26\% \\
D & PCA, AE, VAE & kNN & 9 & T1 halved, T4 doubled (undetermined) & 63\% \\
\bottomrule
\end{tabular}

\vspace{1em}
\begin{tabular}{@{}l>{\raggedright\arraybackslash}p{0.1\textwidth}>{\raggedright\arraybackslash}p{0.12\textwidth}>{\raggedright\arraybackslash}p{0.2\textwidth}>{\raggedright\arraybackslash}p{0.1\textwidth}>{\raggedright\arraybackslash}p{0.13\textwidth}>{\raggedright\arraybackslash}p{0.12\textwidth}@{}}
\multicolumn{7}{@{}l}{\textit{(b) Public ledger}} \\
\toprule
$k$ & P@$k$ leader & FSR leader, equal weights & FSR leader kept (of 17 schemes) & Smallest $\tau_b$ & Random vectors with leader kept & Median $\tau_b$, random \\
\midrule
100 & AE-emb & AE-emb & 16 (kNN when M4 is doubled) & 0.94 & 84\% & 1.00 \\
1,000 & kNN & kNN & 17 & 0.83 & 98\% & 0.94 \\
\bottomrule
\end{tabular}
\end{table}

\clearpage

\bibliographystyle{elsarticle-harv}
\bibliography{refs}

\end{document}


\begin{frontmatter}

\title{Supplementary material for: Machine learning for journal entry testing: A type-aware
evaluation of anomaly detectors under a review budget}

\author[uds,dfki]{Jan Gronewald\corref{cor1}}
\ead{jan.gronewald@extern.uni-saarland.de}
\author[uds,dfki]{Michel Scherer}
\author[leipzig,dfki]{Nijat Mehdiyev}

\cortext[cor1]{Corresponding author.}

\affiliation[uds]{organization={Saarland University},
            addressline={Campus},
            postcode={66123},
            city={Saarbr\"ucken},
            country={Germany}}

\affiliation[dfki]{organization={German Research Center for Artificial Intelligence (DFKI), Institute for Information Systems (IWi)},
            addressline={Campus D3.2},
            postcode={66123},
            city={Saarbr\"ucken},
            country={Germany}}

\affiliation[leipzig]{organization={Leipzig University},
            addressline={Grimmaische Straße 12},
            postcode={04109},
            city={Leipzig},
            country={Germany}}

\end{frontmatter}

This supplement reports the detailed results behind the main text and documents the protocol at the level of implementation. Tables~\ref{tab:s-core}, \ref{tab:s-pertype}, \ref{tab:s-pertype-cont} and \ref{tab:s-sweep} give the full results on the client ledgers, into which typed synthetic anomalies were injected, and Tables~\ref{tab:s-firsthit}, \ref{tab:s-marking}, \ref{tab:s-public-global} and \ref{tab:s-rules} those on the public synthetic ledger. The notation, the objective of DeepSAD, the bootstrap intervals on the public ledger and the sensitivity of fair-share type recall to the type weights are part of the article itself, as Appendices A to D. Section and table numbers without the prefix S refer to the main text.

\section{Protocol details and deviations}\label{sec:s-protocol}

The protocol fixed the injection, the encodings, the detector settings and the measures before the runs, and every departure from it is an implementation detail recorded per cell. Injection follows one random permutation of the postings per seed: the contamination rate times the ledger size, rounded, gives the number of injected postings, T2 to T5 receive fixed shares and T1 the remainder, and no posting carries two types. Cells of one ledger and seed share their leading postings, which reduces the variance between contamination levels. T1 and T2 set the modified z-score of the amount and the payment period to a multiple of the ledger's 99.9th percentile, T3 and T4 draw a fresh identifier per posting, and T5 draws account pairs whose accounts each occur often enough but never together.

Table~\ref{tab:s-deviations} lists the deviations, none of which changes an injection definition, a measure or a detector setting on ledgers C, A and B. On ledger A at the easy level and 3 percent contamination the pool of account pairs is too small for T5, so some pairs are reused, as the protocol permits, and on ledger D the three networks train for fewer epochs with a larger batch (Section 4.3), which the protocol grants for AE-emb and a runtime probe extended to AE and VAE. The PCA score is PyOD's own, a sum of distances to the retained eigenvectors, each divided by its explained-variance ratio, rather than a literal reconstruction error, and it is computed with a memory-saving eigensolver; ECOD is computed in column blocks with the library's scores. T5 is classified as Type V in the typology of \citet{foorthuis2021nature}, not as the Type VI that the protocol named. The optional contextual extreme-amount type and a random-label sensitivity analysis were not run, and the sweep covers ledgers C, A and B only. Because per-seed values were not retained, the Friedman test uses seed-averaged sweep cells and the client-ledger type-aware measures use seed-mean per-type recalls printed to two decimals (Section 3.5).

\section{Field mapping of the public ledger}\label{sec:s-mapping}

Table~\ref{tab:s-mapping} maps every feature field of the client ledgers to a field of the public ledger, so that the same encodings and detector settings apply to both sources. Three fields had to be derived or added: the contra account is the account of the largest-amount line on the opposite side of the entry, with 21 levels; the payment period becomes the entry lag in days; and the sine and cosine of the hour and of the weekday of entry are added, as in the preprocessing of the dataset paper \citep{gronewald2026generating}. Rule attributes, labels, markings, identifiers, free text and account names never enter a feature set, and we checked that no rule attribute or label reaches an encoding. All detector settings on the public ledger are those of ledger D.

\section{Cross-machine reproduction}\label{sec:s-reproduction}

All client-ledger results come from a complete run of the benchmark by a second analyst on another machine, and an earlier partial run, on the machine of Table~\ref{tab:s-runtime}, allows a comparison cell by cell at the printed precision. Ledger D and one sweep cell of ledger B are left out of the comparison, because the two runs averaged different numbers of seeds there. The data audits of both runs agree in row counts, numbers of distinct values, percentiles and account-pair pools; the dataset hashes were not compared byte by byte, but the identical non-neural results are strong indirect evidence that the injected ledgers are the same.

Of 648 compared cells (core-grid means and standard deviations, per-type recalls, sweep and supervised reference), IF, LOF, kNN, HBOS, ECOD, PCA and the LightGBM reference agree in all 435 of theirs, whereas AE, VAE and AE-emb agree in fewer than half of theirs. AE and VAE move by at most 0.06, and AE-emb by up to 0.09 in a core-grid mean and 0.10 in a per-type recall; the probable cause is that network training, reproducible on one machine, is not bitwise reproducible across processors and linear algebra libraries. The type-aware conclusions hold nevertheless: computed from the earlier run's per-type recalls, $\mathrm{FSR@}100$ ranks the same detector highest on ledgers C, A and B, and PCA, which reproduces exactly, ranks above kNN by FSR on ledgers C and A in both runs.

\section{Runtimes}\label{sec:s-runtimes}

Table~\ref{tab:s-runtime} reports wall-clock seconds per fit; the client-ledger times come from the earlier run, whose hardware is documented. Every fit on ledgers C, A and B takes under three minutes, whereas on ledger D LOF and kNN take about 17 minutes each and AE-emb about 86, and on the public ledger LOF and kNN take about eight minutes and every other detector less. The networks train for fewer epochs on ledger D and on the public ledger, so those times are not comparable with the times on ledgers C, A and B.

\begin{table}[tbp]
\caption{Core grid on the client ledgers (contamination 1.5 percent, medium difficulty): precision at budgets of 25, 50 and 100 postings, average precision and ROC-AUC, mean (standard deviation) over five seeds. Bold marks the best unsupervised detector per column within a ledger, on the printed values. The supervised reference is available as a mean for P@25, P@100 and AP only (n/a: not recorded).}
\label{tab:s-core}
\centering
\scriptsize
\setlength{\tabcolsep}{4pt}
\begin{tabular}{@{}lrrrrr@{}}
\toprule
Detector & P@25 & P@50 & P@100 & AP & ROC-AUC \\
\midrule
\multicolumn{6}{@{}l}{\textit{(a) Ledger C: 3,461 postings, 52 injected anomalies, 5 seeds}} \\
IF & 0.04 (0.03) & 0.04 (0.02) & 0.04 (0.01) & 0.04 (0.01) & 0.67 (0.03) \\
LOF & 0.18 (0.07) & 0.11 (0.05) & 0.06 (0.03) & 0.11 (0.02) & 0.83 (0.04) \\
kNN & \textbf{0.65 (0.02)} & \textbf{0.64 (0.02)} & \textbf{0.40 (0.02)} & \textbf{0.52 (0.02)} & \textbf{0.98 (0.00)} \\
HBOS & 0.02 (0.04) & 0.02 (0.03) & 0.03 (0.01) & 0.03 (0.01) & 0.66 (0.03) \\
ECOD & 0.02 (0.04) & 0.08 (0.02) & 0.08 (0.01) & 0.05 (0.01) & 0.79 (0.02) \\
PCA & 0.18 (0.04) & 0.13 (0.02) & 0.32 (0.02) & 0.23 (0.02) & 0.89 (0.01) \\
AE & 0.16 (0.00) & 0.29 (0.10) & 0.38 (0.01) & 0.31 (0.03) & 0.91 (0.02) \\
VAE & 0.38 (0.31) & 0.53 (0.20) & 0.39 (0.02) & 0.44 (0.13) & 0.93 (0.02) \\
AE-emb & 0.12 (0.05) & 0.13 (0.02) & 0.12 (0.02) & 0.09 (0.02) & 0.87 (0.04) \\
\textit{LightGBM reference} & 0.88 & n/a & 0.40 & 0.67 & n/a \\
\midrule
\multicolumn{6}{@{}l}{\textit{(b) Ledger A: 5,080 postings, 76 injected anomalies, 5 seeds}} \\
IF & 0.06 (0.04) & 0.05 (0.02) & 0.03 (0.02) & 0.03 (0.01) & 0.74 (0.02) \\
LOF & 0.02 (0.02) & 0.01 (0.01) & 0.01 (0.01) & 0.04 (0.01) & 0.75 (0.04) \\
kNN & \textbf{0.44 (0.06)} & \textbf{0.39 (0.06)} & \textbf{0.29 (0.04)} & \textbf{0.25 (0.04)} & 0.90 (0.02) \\
HBOS & 0.02 (0.02) & 0.02 (0.02) & 0.02 (0.02) & 0.03 (0.00) & 0.69 (0.01) \\
ECOD & 0.04 (0.06) & 0.03 (0.03) & 0.03 (0.01) & 0.04 (0.00) & 0.78 (0.01) \\
PCA & 0.24 (0.00) & 0.12 (0.01) & 0.15 (0.02) & 0.09 (0.01) & 0.87 (0.01) \\
AE & 0.24 (0.00) & 0.17 (0.06) & 0.22 (0.06) & 0.19 (0.02) & \textbf{0.93 (0.01)} \\
VAE & 0.25 (0.02) & 0.16 (0.05) & 0.16 (0.02) & 0.16 (0.01) & 0.91 (0.01) \\
AE-emb & 0.35 (0.08) & 0.29 (0.02) & 0.19 (0.01) & 0.16 (0.03) & 0.88 (0.02) \\
\textit{LightGBM reference} & 0.98 & n/a & 0.56 & 0.73 & n/a \\
\midrule
\multicolumn{6}{@{}l}{\textit{(c) Ledger B: 17,505 postings, 263 injected anomalies, 5 seeds}} \\
IF & 0.10 (0.07) & 0.06 (0.04) & 0.04 (0.04) & 0.03 (0.01) & 0.65 (0.02) \\
LOF & 0.00 (0.00) & 0.00 (0.00) & 0.00 (0.00) & 0.02 (0.00) & 0.54 (0.01) \\
kNN & 0.26 (0.05) & 0.35 (0.05) & \textbf{0.42 (0.03)} & \textbf{0.36 (0.02)} & \textbf{0.96 (0.01)} \\
HBOS & 0.05 (0.07) & 0.04 (0.05) & 0.03 (0.03) & 0.02 (0.00) & 0.65 (0.02) \\
ECOD & 0.06 (0.07) & 0.07 (0.05) & 0.08 (0.03) & 0.05 (0.00) & 0.82 (0.01) \\
PCA & 0.72 (0.00) & \textbf{0.42 (0.01)} & 0.22 (0.01) & 0.12 (0.00) & 0.86 (0.00) \\
AE & \textbf{0.76 (0.00)} & \textbf{0.42 (0.01)} & 0.23 (0.00) & 0.18 (0.01) & 0.90 (0.00) \\
VAE & \textbf{0.76 (0.03)} & \textbf{0.42 (0.01)} & 0.24 (0.03) & 0.19 (0.01) & 0.89 (0.01) \\
AE-emb & 0.32 (0.23) & 0.27 (0.13) & 0.23 (0.07) & 0.14 (0.03) & 0.89 (0.01) \\
\textit{LightGBM reference} & 0.98 & n/a & 0.97 & 0.86 & n/a \\
\midrule
\multicolumn{6}{@{}l}{\textit{(d) Ledger D: 279,975 postings, 4,200 injected anomalies, 5 seeds}} \\
IF & 0.11 (0.11) & 0.10 (0.09) & 0.11 (0.07) & 0.04 (0.00) & 0.80 (0.01) \\
LOF & 0.45 (0.11) & 0.23 (0.05) & 0.11 (0.03) & 0.06 (0.00) & 0.64 (0.01) \\
kNN & 0.21 (0.05) & 0.27 (0.05) & 0.34 (0.04) & 0.22 (0.00) & 0.92 (0.00) \\
HBOS & 0.02 (0.02) & 0.04 (0.02) & 0.04 (0.01) & 0.04 (0.00) & 0.80 (0.00) \\
ECOD & 0.07 (0.05) & 0.05 (0.03) & 0.05 (0.01) & 0.08 (0.00) & 0.92 (0.00) \\
PCA & 0.92 (0.00) & \textbf{0.96 (0.00)} & \textbf{0.98 (0.00)} & 0.29 (0.00) & 0.95 (0.00) \\
AE & \textbf{0.93 (0.02)} & \textbf{0.96 (0.01)} & \textbf{0.98 (0.00)} & \textbf{0.35 (0.01)} & \textbf{0.97 (0.00)} \\
VAE & \textbf{0.93 (0.02)} & \textbf{0.96 (0.01)} & \textbf{0.98 (0.00)} & 0.30 (0.01) & 0.95 (0.00) \\
AE-emb & 0.41 (0.13) & 0.33 (0.10) & 0.27 (0.11) & 0.13 (0.01) & 0.88 (0.02) \\
\textit{LightGBM reference} & 0.97 & n/a & 0.99 & 0.97 & n/a \\
\bottomrule
\end{tabular}
\end{table}

\begin{table}[tbp]
\caption{Protocol deviations and implementation notes. None changes an injection definition, a metric or a detector hyperparameter on ledgers C, A and B, and each is recorded per cell in the harness output.}
\label{tab:s-deviations}
\centering
\footnotesize
\begin{tabular}{@{}>{\raggedright\arraybackslash}p{0.2\linewidth}>{\raggedright\arraybackslash}p{0.76\linewidth}@{}}
\toprule
Area & Deviation or implementation note \\
\midrule
Injection & Ledger A, easy level, $c = 3$\,\%: the pool holds 19 valid account pairs for the 36 T5 rows the cell needs, so 17 rows reuse a pair. The protocol permits drawing with replacement once the pool is exhausted; the reuse is recorded per cell. \\
Typology & The protocol called T5 a multidimensional mixed-data anomaly (Foorthuis Type VI). T5 changes two categorical fields only, so it is classified as Type V, subtype Va. \\
PCA & The score is PyOD's own: the sum of the distances between the standardised row and the retained eigenvectors, each divided by its explained-variance ratio, not a literal reconstruction error. The covariance eigensolver replaces the full singular value decomposition (same components, scores agree to $4\times10^{-4}$ relative or better, identical P@$k$ and AP). \\
ECOD & PyOD's algorithm recomputed column chunk by column chunk, because the library implementation exhausts memory on ledger D; maximum difference to the library $10^{-14}$ on ledgers C, A and B. Used on every ledger. \\
Networks on ledger D & AE, VAE and AE-emb train with batch 256 for 30 epochs instead of batch 64 for 100; the protocol grants this for AE-emb, and a runtime probe extended it to the two PyOD networks. \\
Execution on ledger D & One thread per process for the linear algebra libraries and PyTorch; every detector runs in a checkpointing worker process; LOF and kNN are fitted on all 279,975 rows, the subsampling rule being implemented but never triggered. Scores are unaffected. \\
Not run & The optional contextual extreme-amount type and the random-label sensitivity analysis were not run. The contamination and difficulty sweep covers ledgers C, A and B; every ledger D result is at $c = 1.5$\,\% and medium difficulty. \\
Public ledger & Networks at the ledger D setting; stochastic detectors on seeds 0 to 2; sine and cosine of the hour and weekday of the entry timestamp added to the encodings; posting lines are scored and journal entries evaluated (maximum over the lines); budget $k = 1{,}000$ added; contra account derived; rule attributes excluded; cross-fitting folds of the supervised reference grouped by entry. \\
Analysis & The Friedman test uses the 27 seed-averaged sweep cells of ledgers C, A and B, because per-seed values of the sweep were not retained. The type-aware measures on the client ledgers are computed from seed-mean per-type recalls at $k = 100$ printed to two decimals and are reported with the bounds this rounding implies. \\
Added in revision & A no-feedback control of the DeepSAD review protocol on the public ledger, seeds 0 to 2: the same initialisation, rounds, seeds, optimiser and retraining as the protocol run, with the revealed labels withheld from the retraining. It was run after the frozen experiments, is reported as a control only and changes no protocol result. \\
\bottomrule
\end{tabular}
\end{table}

\begin{table}[tbp]
\caption{Mapping of the benchmark onto the public ledger, field by field. The protocol expects the eleven columns of the client ledgers; the table states which field of the public ledger takes each role, what had to be derived and what was added because the public ledger records timestamps. The rule attributes supplied with the dataset are excluded from every feature set, as its authors recommend. All detector settings are those of ledger D.}
\label{tab:s-mapping}
\centering
\footnotesize
\begin{tabular}{@{}>{\raggedright\arraybackslash}p{0.2\linewidth}>{\raggedright\arraybackslash}p{0.3\linewidth}>{\raggedright\arraybackslash}p{0.42\linewidth}@{}}
\toprule
Role & Client ledgers & Public ledger \\
\midrule
Scored row & one posting & one posting line \\
Evaluated unit & one posting & one journal entry; its score is the maximum over its lines, and labels are constant within an entry \\
Account & \texttt{Kontonummer} & \texttt{gl\_account} \\
Contra account & \texttt{Gegenkonto} & derived: the account of the largest-amount line on the opposite side of the same entry, 21 levels \\
Amount & \texttt{Umsatz\_mod\_z}, the modified z-score supplied with the data & \texttt{amount}; $\log(1+x)$ and standardised in the encodings \\
Payment period & \texttt{Zahlungszeitraum} & entry lag in days, \texttt{entered\_date} minus \texttt{document\_date}; used in the encodings \\
Currency & \texttt{W\"ahrung} & debit or credit flag (\texttt{cd\_flag}) \\
Tax code & \texttt{Steuersatz} & \texttt{tax\_rate} \\
User & \texttt{Benutzererkennung} & \texttt{user} \\
Round amount flag & \texttt{Umsatz\_Smooth} & amount a multiple of EUR 10 \\
Document date outside the fiscal year & \texttt{Abweichung Belegdatum zum WJ} & document date in a different year than the posting date (constant 0 here) \\
Posting text present & \texttt{Buchungstext nicht Leer} & free-text field non-empty (constant 1 here) \\
Time of entry & no counterpart: the client ledgers carry no timestamps & added: sine and cosine of the hour and of the weekday of the entry timestamp, as in the preprocessing of the dataset paper \\
Never used as a feature & the posting key \texttt{BUCHUNGS\_GUID} & the precomputed rule attributes \texttt{promptly}, \texttt{weekend}, \texttt{nwh}, \texttt{top\_n} and \texttt{high\_cash}, the \texttt{label} and \texttt{marking}, \texttt{posting\_id} and \texttt{ref\_id}, the free text and the account name \\
\bottomrule
\end{tabular}
\end{table}

\begin{table}[tbp]
\caption{Public ledger, first-hit rank per marking: the number of journal entries an auditor reviews until the first anomaly of the marking appears. All is the rank at which every marking has appeared at least once (the largest first-hit rank). Stochastic detectors and the supervised reference: mean over three seeds, with the smallest and largest value over the seeds in italics; the mean of All is taken over the per-seed maxima.}
\label{tab:s-firsthit}
\centering
\footnotesize
\setlength{\tabcolsep}{4pt}
\begin{tabular}{@{}lrrrrrrrrr@{}}
\toprule
Detector & M1 & M2 & M3 & M4 & M5 & M6 & M7 & M8 & All \\
\midrule
IF & 1 & 51 & 626 & 309 & 46 & 217 & 3,415 & 774 & 3,415 \\
\quad\textit{min} & \textit{1} & \textit{37} & \textit{327} & \textit{74} & \textit{12} & \textit{140} & \textit{2,157} & \textit{661} & \textit{2,157} \\
\quad\textit{max} & \textit{1} & \textit{62} & \textit{1,108} & \textit{654} & \textit{92} & \textit{279} & \textit{4,486} & \textit{944} & \textit{4,486} \\
LOF & 239 & 3,116 & 472 & 11,107 & 192 & 3,098 & 1 & 6,895 & 11,107 \\
kNN & 43 & 185 & 824 & 15 & 70 & 1,001 & 53 & 4,641 & 4,641 \\
HBOS & 1 & 429 & 1,386 & 661 & 456 & 479 & 30 & 703 & 1,386 \\
ECOD & 4 & 602 & 1,063 & 721 & 395 & 455 & 30 & 1,103 & 1,103 \\
PCA & 8 & 890 & 1,070 & 643 & 737 & 4,592 & 7 & 5,529 & 5,529 \\
AE & 87 & 1,228 & 1,394 & 364 & 654 & 2,456 & 7 & 6,493 & 6,493 \\
\quad\textit{min} & \textit{78} & \textit{693} & \textit{1,028} & \textit{285} & \textit{499} & \textit{888} & \textit{7} & \textit{3,923} & \textit{3,923} \\
\quad\textit{max} & \textit{104} & \textit{1,914} & \textit{2,016} & \textit{404} & \textit{736} & \textit{4,623} & \textit{7} & \textit{8,704} & \textit{8,704} \\
VAE & 136 & 1,313 & 1,327 & 364 & 726 & 2,403 & 7 & 6,933 & 6,933 \\
\quad\textit{min} & \textit{30} & \textit{886} & \textit{1,027} & \textit{285} & \textit{721} & \textit{1,261} & \textit{7} & \textit{5,951} & \textit{5,951} \\
\quad\textit{max} & \textit{207} & \textit{2,007} & \textit{1,526} & \textit{404} & \textit{736} & \textit{4,601} & \textit{7} & \textit{7,484} & \textit{7,484} \\
AE-emb & 67 & 1,894 & 2,154 & 184 & 79 & 2,801 & 6 & 5,198 & 5,417 \\
\quad\textit{min} & \textit{8} & \textit{102} & \textit{1,476} & \textit{6} & \textit{13} & \textit{1,161} & \textit{1} & \textit{1,301} & \textit{1,956} \\
\quad\textit{max} & \textit{177} & \textit{4,915} & \textit{3,122} & \textit{296} & \textit{191} & \textit{5,286} & \textit{16} & \textit{7,494} & \textit{7,494} \\
\addlinespace
\textit{LightGBM reference} & 5 & 35 & 587 & 55 & 15 & 124 & 24 & 4,900 & 4,900 \\
\quad\textit{min} & \textit{1} & \textit{3} & \textit{504} & \textit{7} & \textit{6} & \textit{53} & \textit{16} & \textit{2,543} & \textit{2,543} \\
\quad\textit{max} & \textit{12} & \textit{98} & \textit{675} & \textit{100} & \textit{21} & \textit{234} & \textit{38} & \textit{7,622} & \textit{7,622} \\
\bottomrule
\end{tabular}
\end{table}

\begin{table}[tbp]
\caption{Public ledger, recall per marking within the first 100 and the first 1,000 journal entries, mean over seeds for the stochastic detectors. Bold marks the best unsupervised detector per column (none where all are zero). M1 high amount, M2 high cash disbursement, M3 delayed entry, M4 very delayed entry, M5 non-working hours, M6 weekend, M7 shadow bank account, M8 cross-linked clearing; $n_t$ is the number of anomalous entries per marking.}
\label{tab:s-marking}
\centering
\footnotesize
\begin{tabular}{@{}lrrrrrrrr@{}}
\toprule
Detector & M1 & M2 & M3 & M4 & M5 & M6 & M7 & M8 \\
\textit{$n_t$} & \textit{120} & \textit{60} & \textit{120} & \textit{120} & \textit{100} & \textit{250} & \textit{60} & \textit{20} \\
\midrule
\multicolumn{9}{@{}l}{\textit{(a) Recall within the first 100 entries}} \\
IF & 0.02 & \textbf{0.04} & 0.00 & 0.00 & 0.02 & 0.00 & 0.00 & 0.00 \\
LOF & 0.00 & 0.00 & 0.00 & 0.00 & 0.00 & 0.00 & 0.18 & 0.00 \\
kNN & 0.02 & 0.00 & 0.00 & \textbf{0.07} & \textbf{0.03} & 0.00 & 0.07 & 0.00 \\
HBOS & 0.02 & 0.00 & 0.00 & 0.00 & 0.00 & 0.00 & 0.03 & 0.00 \\
ECOD & 0.02 & 0.00 & 0.00 & 0.00 & 0.00 & 0.00 & 0.03 & 0.00 \\
PCA & 0.02 & 0.00 & 0.00 & 0.00 & 0.00 & 0.00 & 0.02 & 0.00 \\
AE & 0.01 & 0.00 & 0.00 & 0.00 & 0.00 & 0.00 & 0.02 & 0.00 \\
VAE & 0.01 & 0.00 & 0.00 & 0.00 & 0.00 & 0.00 & 0.02 & 0.00 \\
AE-emb & \textbf{0.20} & 0.00 & 0.00 & 0.01 & 0.02 & 0.00 & \textbf{0.21} & 0.00 \\
\addlinespace
\textit{LightGBM reference} & 0.19 & 0.07 & 0.00 & 0.03 & 0.16 & 0.00 & 0.59 & 0.00 \\
\midrule
\multicolumn{9}{@{}l}{\textit{(b) Recall within the first 1,000 entries}} \\
IF & 0.08 & 0.24 & \textbf{0.02} & 0.13 & 0.18 & 0.02 & 0.00 & \textbf{0.05} \\
LOF & 0.13 & 0.00 & \textbf{0.02} & 0.00 & 0.09 & 0.00 & 0.78 & 0.00 \\
kNN & 0.07 & 0.08 & 0.01 & \textbf{0.68} & \textbf{0.63} & 0.00 & 0.93 & 0.00 \\
HBOS & 0.13 & \textbf{0.25} & 0.00 & 0.01 & 0.14 & \textbf{0.03} & 0.23 & \textbf{0.05} \\
ECOD & 0.06 & 0.10 & 0.00 & 0.02 & 0.23 & 0.02 & 0.67 & 0.00 \\
PCA & 0.02 & 0.03 & 0.00 & 0.25 & 0.07 & 0.00 & \textbf{1.00} & 0.00 \\
AE & 0.02 & 0.01 & 0.00 & 0.29 & 0.12 & 0.00 & \textbf{1.00} & 0.00 \\
VAE & 0.02 & 0.01 & 0.00 & 0.26 & 0.07 & 0.00 & \textbf{1.00} & 0.00 \\
AE-emb & \textbf{0.47} & 0.02 & 0.00 & 0.08 & 0.29 & 0.00 & 0.98 & 0.00 \\
\addlinespace
\textit{LightGBM reference} & 0.97 & 0.59 & 0.43 & 0.96 & 0.97 & 0.79 & 0.98 & 0.00 \\
\bottomrule
\end{tabular}
\end{table}

\begin{table}[tbp]
\caption{Recall within the first 100 postings per injected type ($R_t$@100) on the core grid of ledgers C and A, mean over five seeds as printed in the result tables, with the fair-share type recall FSR@100 derived from them. $n_t$ is the number of injected rows of type $t$ and the budget ceiling $\min(1, 100/n_t)$ the largest attainable $R_t$@100. Because the per-type recalls are seed means rounded to two decimals, FSR@100 is given as the tightest bounds consistent with the printed values and the printed P@100 (a single value where both bounds agree at the printed precision). Type coverage is not reported, because the seed-mean recalls do not determine it. T1 extreme amount, T2 extreme payment period, T3 unseen contra account, T4 unseen account, T5 unusual account pair.}
\label{tab:s-pertype}
\centering
\footnotesize
\begin{tabular}{@{}lrrrrrr@{}}
\toprule
Detector & T1 & T2 & T3 & T4 & T5 & FSR@100 \\
\midrule
\multicolumn{7}{@{}l}{\textit{(a) Ledger C}} \\
\textit{$n_t$} & \textit{22} & \textit{10} & \textit{4} & \textit{4} & \textit{12} & \\
IF & 0.12 & 0.16 & 0.05 & 0.00 & 0.00 & 0.07 \\
LOF & 0.22 & 0.00 & 0.00 & 0.30 & 0.03 & 0.11 \\
kNN & 1.00 & 1.00 & 0.15 & 0.10 & 0.58 & 0.57 \\
HBOS & 0.06 & 0.04 & 0.05 & 0.10 & 0.03 & 0.06 \\
ECOD & 0.20 & 0.12 & 0.05 & 0.50 & 0.02 & 0.18 \\
PCA & 0.76 & 1.00 & 0.05 & 1.00 & 0.05 & 0.57 \\
AE & 1.00 & 1.00 & 0.00 & 1.00 & 0.18 & 0.64 \\
VAE & 1.00 & 1.00 & 0.00 & 1.00 & 0.27 & 0.65 \\
AE-emb & 0.20 & 0.26 & 0.05 & 0.10 & 0.37 & 0.20 \\
\textit{Budget ceiling} & \textit{1.00} & \textit{1.00} & \textit{1.00} & \textit{1.00} & \textit{1.00} & \\
\midrule
\multicolumn{7}{@{}l}{\textit{(b) Ledger A}} \\
\textit{$n_t$} & \textit{31} & \textit{15} & \textit{6} & \textit{6} & \textit{18} & \\
IF & 0.06 & 0.07 & 0.00 & 0.00 & 0.03 & 0.03 \\
LOF & 0.01 & 0.00 & 0.00 & 0.10 & 0.00 & 0.02 \\
kNN & 0.49 & 0.47 & 0.00 & 0.13 & 0.34 & 0.29 \\
HBOS & 0.03 & 0.00 & 0.00 & 0.03 & 0.03 & 0.02 \\
ECOD & 0.04 & 0.05 & 0.00 & 0.10 & 0.02 & 0.04 \\
PCA & 0.08 & 0.20 & 0.00 & 1.00 & 0.21 & 0.30 \\
AE & 0.09 & 0.56 & 0.00 & 1.00 & 0.26 & 0.38 \\
VAE & 0.05 & 0.25 & 0.00 & 1.00 & 0.26 & 0.31 \\
AE-emb & 0.37 & 0.23 & 0.03 & 0.20 & 0.16 & 0.20 \\
\textit{Budget ceiling} & \textit{1.00} & \textit{1.00} & \textit{1.00} & \textit{1.00} & \textit{1.00} & \\
\bottomrule
\end{tabular}
\end{table}

\begin{table}[tbp]
\caption{Recall within the first 100 postings per injected type, continued from Table~\ref{tab:s-pertype} for ledgers B and D. On ledger D every type holds at least 336 injected rows, so no per-type recall can exceed 0.30.}
\label{tab:s-pertype-cont}
\centering
\footnotesize
\begin{tabular}{@{}lrrrrrr@{}}
\toprule
Detector & T1 & T2 & T3 & T4 & T5 & FSR@100 \\
\midrule
\multicolumn{7}{@{}l}{\textit{(c) Ledger B}} \\
\textit{$n_t$} & \textit{105} & \textit{53} & \textit{21} & \textit{21} & \textit{63} & \\
IF & 0.02 & 0.04 & 0.00 & 0.00 & 0.00 & 0.04 \\
LOF & 0.00 & 0.00 & 0.00 & 0.00 & 0.00 & 0.00 \\
kNN & 0.29 & 0.20 & 0.01 & 0.00 & 0.01 & 0.31 to 0.32 \\
HBOS & 0.02 & 0.01 & 0.00 & 0.02 & 0.01 & 0.03 \\
ECOD & 0.04 & 0.02 & 0.01 & 0.12 & 0.00 & 0.08 \\
PCA & 0.01 & 0.00 & 0.00 & 1.00 & 0.00 & 0.21 \\
AE & 0.01 & 0.01 & 0.00 & 1.00 & 0.01 & 0.22 \\
VAE & 0.01 & 0.00 & 0.00 & 1.00 & 0.03 & 0.23 \\
AE-emb & 0.10 & 0.08 & 0.05 & 0.04 & 0.11 & 0.23 \\
\textit{Budget ceiling} & \textit{0.95} & \textit{1.00} & \textit{1.00} & \textit{1.00} & \textit{1.00} & \\
\midrule
\multicolumn{7}{@{}l}{\textit{(d) Ledger D}} \\
\textit{$n_t$} & \textit{1,680} & \textit{840} & \textit{336} & \textit{336} & \textit{1,008} & \\
IF & 0.00 & 0.00 & 0.00 & 0.00 & 0.00 & 0.106 to 0.114 \\
LOF & 0.00 & 0.00 & 0.00 & 0.03 & 0.00 & 0.106 to 0.114 \\
kNN & 0.01 & 0.01 & 0.00 & 0.00 & 0.00 & 0.284 to 0.344 \\
HBOS & 0.00 & 0.00 & 0.00 & 0.00 & 0.00 & 0.036 to 0.044 \\
ECOD & 0.00 & 0.00 & 0.00 & 0.00 & 0.00 & 0.046 to 0.054 \\
PCA & 0.00 & 0.00 & 0.00 & 0.29 & 0.00 & 0.200 to 0.226 \\
AE & 0.00 & 0.00 & 0.00 & 0.29 & 0.00 & 0.200 to 0.226 \\
VAE & 0.00 & 0.00 & 0.00 & 0.29 & 0.00 & 0.200 to 0.226 \\
AE-emb & 0.01 & 0.01 & 0.01 & 0.00 & 0.01 & 0.266 to 0.274 \\
\textit{Budget ceiling} & \textit{0.06} & \textit{0.12} & \textit{0.30} & \textit{0.30} & \textit{0.10} & \\
\bottomrule
\end{tabular}
\end{table}

\begin{table}[tbp]
\caption{Public ledger, global ranking of all 102,429 journal entries, of which 850 are anomalous. Mean (standard deviation) over seeds 0 to 2 for the stochastic detectors and the supervised reference; the data are fixed, so deterministic detectors have a single run. Bold marks the best unsupervised detector per column on the printed values. The last three columns split the first 100 entries into anomalous entries, normal entries on a rare account (fewer than 1,000 lines) and other normal entries (means over seeds).}
\label{tab:s-public-global}
\centering
\scriptsize
\setlength{\tabcolsep}{4pt}
\begin{tabular}{@{}lrrrr@{}}
\toprule
\multicolumn{5}{@{}l}{\textit{(a) Precision at review budgets}} \\
\addlinespace
Detector & P@25 & P@50 & P@100 & P@1,000 \\
\midrule
IF & 0.09 (0.02) & 0.07 (0.03) & 0.08 (0.04) & 0.067 (0.024) \\
LOF & 0.12 & 0.14 & 0.11 & 0.074 \\
kNN & 0.08 & 0.12 & 0.17 & \textbf{0.215} \\
HBOS & 0.08 & 0.08 & 0.04 & 0.067 \\
ECOD & 0.08 & 0.08 & 0.04 & 0.083 \\
PCA & 0.12 & 0.06 & 0.03 & 0.101 \\
AE & 0.04 (0.00) & 0.02 (0.00) & 0.02 (0.01) & 0.111 (0.020) \\
VAE & 0.04 (0.00) & 0.03 (0.02) & 0.02 (0.01) & 0.100 (0.004) \\
AE-emb & \textbf{0.36 (0.21)} & \textbf{0.39 (0.15)} & \textbf{0.40 (0.23)} & 0.154 (0.020) \\
\addlinespace
\textit{LightGBM reference} & 0.52 (0.11) & 0.70 (0.12) & 0.83 (0.09) & 0.672 (0.003) \\
\bottomrule
\end{tabular}

\bigskip
\begin{tabular}{@{}lrrrrrr@{}}
\toprule
\multicolumn{7}{@{}l}{\textit{(b) Threshold-free metrics and composition of the first 100 entries}} \\
\addlinespace
 & & & & \multicolumn{3}{c}{First 100 entries} \\
\cmidrule(l){5-7}
Detector & R@1,000 & AP & ROC-AUC & anomalous & rare account & other \\
\midrule
IF & 0.078 (0.028) & 0.05 (0.01) & 0.83 (0.01) & 7.7 & 24.0 & 68.3 \\
LOF & 0.087 & 0.03 & 0.67 & 11 & 12 & 77 \\
kNN & \textbf{0.253} & \textbf{0.12} & 0.87 & 17 & 6 & 77 \\
HBOS & 0.079 & 0.05 & 0.85 & 4 & 96 & 0 \\
ECOD & 0.098 & 0.06 & 0.88 & 4 & 96 & 0 \\
PCA & 0.119 & 0.05 & 0.85 & 3 & 97 & 0 \\
AE & 0.130 (0.024) & 0.06 (0.01) & \textbf{0.90 (0.01)} & 2.3 & 97.7 & 0.0 \\
VAE & 0.118 (0.004) & 0.06 (0.00) & \textbf{0.90 (0.01)} & 1.7 & 98.3 & 0.0 \\
AE-emb & 0.182 (0.023) & 0.08 (0.02) & 0.71 (0.04) & 40.0 & 31.3 & 28.7 \\
\addlinespace
\textit{LightGBM reference} & 0.791 (0.004) & 0.74 (0.02) & 0.96 (0.00) & 83.0 & 0.0 & 17.0 \\
\bottomrule
\end{tabular}
\end{table}

\begin{table}[tbp]
\caption{Public ledger, ranking inside the four journal entry testing rule populations. Each population is the entry set that one rule returns; inside it the rule attribute is constant, so the ranking cannot use it. Rule precision is the share of anomalous entries in the population, which is what plain rule matching delivers. Means over seeds for the stochastic detectors; bold marks the best unsupervised detector per column.}
\label{tab:s-rules}
\centering
\footnotesize
\begin{tabular}{@{}lrrrr@{}}
\toprule
 & J1 & J2 & J3 & J4 \\
\midrule
Rule & \makecell[r]{top-n\\amounts} & high cash & \makecell[r]{delayed\\entry} & \makecell[r]{weekend or\\non-working hours} \\
Entries & 2,000 & 402 & 8,207 & 16,493 \\
Anomalies & 196 & 190 & 310 & 415 \\
Rule precision & 0.098 & 0.473 & 0.038 & 0.025 \\
\midrule
\multicolumn{5}{@{}l}{\textit{(a) P@100 inside the population}} \\
IF & 0.40 & 0.53 & 0.06 & 0.08 \\
LOF & 0.42 & 0.84 & 0.01 & 0.12 \\
kNN & 0.41 & \textbf{0.87} & 0.16 & 0.21 \\
HBOS & 0.41 & 0.56 & 0.13 & 0.16 \\
ECOD & 0.27 & 0.60 & 0.17 & 0.22 \\
PCA & 0.16 & 0.40 & 0.19 & 0.22 \\
AE & 0.23 & 0.74 & 0.17 & 0.22 \\
VAE & 0.26 & 0.74 & 0.18 & \textbf{0.23} \\
AE-emb & \textbf{0.63} & 0.83 & \textbf{0.22} & 0.22 \\
\addlinespace
\textit{LightGBM reference} & 0.96 & 0.93 & 0.87 & 0.92 \\
\midrule
\multicolumn{5}{@{}l}{\textit{(b) Average precision inside the population}} \\
IF & \textbf{0.435} & 0.576 & 0.086 & 0.065 \\
LOF & 0.244 & 0.773 & 0.032 & 0.044 \\
kNN & 0.372 & \textbf{0.774} & 0.126 & 0.085 \\
HBOS & 0.419 & 0.603 & 0.097 & 0.082 \\
ECOD & 0.278 & 0.617 & 0.099 & 0.107 \\
PCA & 0.369 & 0.485 & 0.122 & 0.112 \\
AE & 0.287 & 0.686 & \textbf{0.139} & \textbf{0.144} \\
VAE & 0.333 & 0.680 & 0.135 & 0.135 \\
AE-emb & 0.404 & 0.768 & 0.098 & 0.082 \\
\addlinespace
\textit{LightGBM reference} & 0.893 & 0.834 & 0.687 & 0.761 \\
\bottomrule
\end{tabular}
\end{table}

\begin{table}[tbp]
\caption{Wall-clock seconds for one detector fit (2 CPU cores, 7 GB RAM, no GPU). Client ledgers: core-grid cells of the first execution of the benchmark, the one whose hardware is documented, mean over 5 seeds on ledgers C, A and B and over 3 seeds on ledger D. Public ledger: mean over the runs of a detector (three seeds for the stochastic detectors and the supervised reference), on two CPU cores. On ledger D and on the public ledger the three networks train for 30 epochs with batch 256 instead of 100 epochs with batch 64, so their times there are not comparable with ledgers C, A and B. n/a: not timed.}
\label{tab:s-runtime}
\centering
\footnotesize
\begin{tabular}{@{}lrrrrr@{}}
\toprule
 & \multicolumn{4}{c}{Client ledgers} & Public \\
\cmidrule(lr){2-5}
Detector & C & A & B & D & ledger \\
\midrule
IF & 14.1 & 14.7 & 0.6 & 9.8 & 5.1 \\
LOF & 0.1 & 0.1 & 1.5 & 1,022.6 & 478.2 \\
kNN & 0.0 & 0.1 & 1.5 & 1,029.0 & 467.1 \\
HBOS & 1.3 & 1.3 & 0.2 & 10.5 & 4.6 \\
ECOD & 0.1 & 0.1 & 0.4 & 25.1 & 4.8 \\
PCA & 0.0 & 0.0 & 0.2 & 16.7 & 1.2 \\
AE & 24.6 & 36.9 & 144.4 & 282.8 & 297.9 \\
VAE & 18.6 & 29.6 & 127.2 & 286.9 & 268.4 \\
AE-emb & 20.2 & 37.6 & 150.4 & 5,179.4 & 182.1 \\
\addlinespace
\textit{LightGBM reference} & n/a & n/a & n/a & n/a & 40.3 \\
\bottomrule
\end{tabular}
\end{table}

\begin{table}[tbp]
\caption{Contamination and difficulty sweep on ledgers C, A and B: P@100 for every combination of the contamination $c$ and the difficulty level, mean over five seeds. The row Maximum is the attainable P@100, $\min(1, n/100)$ for $n$ injected anomalies; at $c = 0.5$ percent every ledger holds fewer than 100 anomalies. The 27 cells are the input of the rank analysis in Section 5.1, which uses the seed-averaged cells because per-seed values were not retained. Bold marks the highest P@100 per column within a ledger.}
\label{tab:s-sweep}
\centering
\footnotesize
\begin{tabular}{@{}lrrrrrrrrr@{}}
\toprule
 & \multicolumn{3}{c}{$c = 0.5$\,\%} & \multicolumn{3}{c}{$c = 1.5$\,\%} & \multicolumn{3}{c}{$c = 3$\,\%} \\
\cmidrule(lr){2-4}\cmidrule(lr){5-7}\cmidrule(lr){8-10}
Detector & easy & medium & hard & easy & medium & hard & easy & medium & hard \\
\midrule
\multicolumn{10}{@{}l}{\textit{(a) Ledger C}} \\
IF & 0.02 & 0.01 & 0.01 & 0.04 & 0.04 & 0.04 & 0.09 & 0.09 & 0.08 \\
LOF & 0.04 & 0.03 & 0.02 & 0.11 & 0.06 & 0.03 & 0.07 & 0.04 & 0.02 \\
kNN & \textbf{0.15} & \textbf{0.14} & \textbf{0.14} & \textbf{0.43} & \textbf{0.40} & \textbf{0.37} & \textbf{0.75} & \textbf{0.69} & \textbf{0.64} \\
HBOS & 0.01 & 0.01 & 0.01 & 0.03 & 0.03 & 0.03 & 0.05 & 0.06 & 0.06 \\
ECOD & 0.03 & 0.03 & 0.02 & 0.08 & 0.08 & 0.07 & 0.10 & 0.11 & 0.09 \\
PCA & 0.12 & 0.12 & 0.10 & 0.34 & 0.32 & 0.16 & 0.34 & 0.25 & 0.17 \\
AE & 0.13 & 0.13 & 0.12 & 0.37 & 0.38 & 0.31 & 0.41 & 0.41 & 0.25 \\
VAE & 0.13 & 0.13 & 0.13 & 0.38 & 0.39 & 0.32 & 0.45 & 0.36 & 0.27 \\
AE-emb & 0.08 & 0.06 & 0.07 & 0.12 & 0.12 & 0.16 & 0.19 & 0.20 & 0.28 \\
\textit{Maximum} & \textit{0.17} & \textit{0.17} & \textit{0.17} & \textit{0.52} & \textit{0.52} & \textit{0.52} & \textit{1.00} & \textit{1.00} & \textit{1.00} \\
\midrule
\multicolumn{10}{@{}l}{\textit{(b) Ledger A}} \\
IF & 0.02 & 0.01 & 0.01 & 0.04 & 0.03 & 0.02 & 0.06 & 0.06 & 0.04 \\
LOF & 0.00 & 0.00 & 0.00 & 0.03 & 0.01 & 0.02 & 0.01 & 0.01 & 0.02 \\
kNN & \textbf{0.22} & \textbf{0.17} & \textbf{0.13} & \textbf{0.42} & \textbf{0.29} & 0.23 & \textbf{0.63} & \textbf{0.50} & \textbf{0.39} \\
HBOS & 0.00 & 0.00 & 0.00 & 0.03 & 0.02 & 0.01 & 0.05 & 0.05 & 0.04 \\
ECOD & 0.01 & 0.01 & 0.01 & 0.03 & 0.03 & 0.02 & 0.06 & 0.06 & 0.04 \\
PCA & 0.17 & 0.07 & 0.04 & 0.23 & 0.15 & 0.11 & 0.21 & 0.24 & 0.19 \\
AE & 0.17 & \textbf{0.17} & 0.04 & 0.28 & 0.22 & 0.13 & 0.22 & 0.26 & 0.22 \\
VAE & 0.17 & 0.14 & 0.04 & 0.22 & 0.16 & 0.12 & 0.22 & 0.27 & 0.21 \\
AE-emb & 0.06 & 0.08 & 0.11 & 0.14 & 0.19 & \textbf{0.24} & 0.25 & 0.26 & 0.26 \\
\textit{Maximum} & \textit{0.25} & \textit{0.25} & \textit{0.25} & \textit{0.76} & \textit{0.76} & \textit{0.76} & \textit{1.00} & \textit{1.00} & \textit{1.00} \\
\midrule
\multicolumn{10}{@{}l}{\textit{(c) Ledger B}} \\
IF & 0.01 & 0.01 & 0.01 & 0.04 & 0.04 & 0.03 & 0.09 & 0.10 & 0.09 \\
LOF & 0.00 & 0.00 & 0.00 & 0.00 & 0.00 & 0.00 & 0.00 & 0.00 & 0.00 \\
kNN & \textbf{0.50} & \textbf{0.42} & \textbf{0.32} & \textbf{0.73} & \textbf{0.42} & \textbf{0.36} & \textbf{0.83} & \textbf{0.53} & 0.47 \\
HBOS & 0.00 & 0.00 & 0.00 & 0.03 & 0.03 & 0.03 & 0.07 & 0.07 & 0.05 \\
ECOD & 0.03 & 0.03 & 0.02 & 0.09 & 0.08 & 0.08 & 0.14 & 0.14 & 0.14 \\
PCA & 0.07 & 0.07 & 0.09 & 0.22 & 0.22 & 0.27 & 0.43 & 0.43 & 0.47 \\
AE & 0.08 & 0.07 & 0.09 & 0.22 & 0.23 & 0.30 & 0.43 & 0.43 & 0.51 \\
VAE & 0.22 & 0.08 & 0.10 & 0.23 & 0.24 & 0.27 & 0.43 & 0.43 & \textbf{0.52} \\
AE-emb & 0.12 & 0.08 & 0.08 & 0.22 & 0.23 & 0.19 & 0.32 & 0.39 & 0.29 \\
\textit{Maximum} & \textit{0.88} & \textit{0.88} & \textit{0.88} & \textit{1.00} & \textit{1.00} & \textit{1.00} & \textit{1.00} & \textit{1.00} & \textit{1.00} \\
\bottomrule
\end{tabular}
\end{table}

\clearpage
\bibliographystyle{elsarticle-harv}
\bibliography{refs.bib}